\documentclass{article}

\usepackage[margin=1in]{geometry}
\usepackage[utf8]{inputenc} %
\usepackage[T1]{fontenc}    %
\usepackage{url}            %
\usepackage{booktabs}       %
\usepackage{amsfonts}       %
\usepackage{amssymb}        %
\usepackage{nicefrac}       %
\usepackage{microtype}      %
\usepackage{xcolor}         %

\usepackage[unicode=true,
 bookmarks=false,
 breaklinks=false,pdfborder={0 0 1},colorlinks=false]{hyperref}
\hypersetup{
colorlinks,citecolor=blue,filecolor=blue,linkcolor=blue,urlcolor=blue}

\usepackage{amsthm}
\usepackage{cite}  
\usepackage{comment}
\usepackage{natbib}
\usepackage{booktabs,mathtools}

\usepackage{graphicx}
\usepackage[linesnumbered,ruled,vlined]{algorithm2e}
\usepackage{algorithmic}

\usepackage{float}
\usepackage{multirow}
\usepackage{cancel}
\usepackage{enumitem}
\setlist[itemize,enumerate]{leftmargin=*}

\usepackage{subcaption}
\usepackage{setspace}

\usepackage{dsfont}
\usepackage{color}
\definecolor{yjc}{RGB}{125,0,0}
\definecolor{jiw}{RGB}{10,148,15}

\newcommand{\orange}[1]{\textcolor{orange}{#1}}
\newcommand{\magenta}[1]{\textcolor{magenta}{#1}}
\newcommand{\yjc}[1]{{{{\magenta{{[Yuejie --- #1]}}}}}}
\let\yc\yjc %
\newcommand{\GJ}[1]{{{{\orange{{[GJ: #1]}}}}}}
\newcommand{\jiw}[1]{\textcolor{jiw}{[JIW: #1]}}

\newcommand{\rev}[2]{{\color{red}\textbf{[#1]}~#2}}
\newcommand{\lfix}[2]{{\color{violet}\textbf{[Lemma3 fix-#1]}~#2}}
\newif\ifshowrev
\showrevfalse
\makeatletter
\DeclareRobustCommand{\revb}[1]{\ifshowrev\expandafter\@firstoftwo\else\expandafter\@secondoftwo\fi{\relax\ifmmode\boldsymbol{#1}\else{\boldmath\bfseries #1}\fi}{#1}}
\DeclareRobustCommand{\bred}[1]{\ifshowrev\expandafter\@firstoftwo\else\expandafter\@secondoftwo\fi{\textcolor{red}{\revb{#1}}}{#1}}
\makeatother

\ifshowrev\else
  \renewcommand{\rev}[2]{}\renewcommand{\lfix}[2]{}\renewcommand{\jiw}[1]{}
  \renewcommand{\yc}[1]
  {}\renewcommand{\GJ}[1]{}
  \newcommand{\wfix}[2]{}
\fi

\DeclareMathOperator*{\argmax}{arg\,max}

\allowdisplaybreaks
\renewcommand{\tilde}{\widetilde}

\newcommand{\pfefvi}{{\sf PF-EFVI}\xspace}
\newcommand{\pfucbvi}{{\sf PF-UCBVI}\xspace}

\renewcommand{\hat}{\widehat}

\DeclareMathOperator{\ind}{\mathds{1}}  %

\newcommand{\defn}{\coloneqq}

\newcommand{\mprob}{\mathbb{P}}
\newcommand{\mexp}{\mathbb{E}}

\newcommand{\Var}{\mathsf{Var}}

\usepackage{scalerel,stackengine}
\stackMath
\newcommand\reallywidehat[1]{%
\savestack{\tmpbox}{\stretchto{%
  \scaleto{%
    \scalerel*[\widthof{\ensuremath{#1}}]{\kern-.6pt\bigwedge\kern-.6pt}%
    {\rule[-\textheight/2]{1ex}{\textheight}}%
  }{\textheight}%
}{0.5ex}}%
\stackon[1pt]{#1}{\tmpbox}%
}
\newcommand\reallywidecheck[1]{%
\savestack{\tmpbox}{\stretchto{%
  \scaleto{%
    \scalerel*[\widthof{\ensuremath{#1}}]{\kern-.6pt\bigwedge\kern-.6pt}%
    {\rule[-\textheight/2]{1ex}{\textheight}}%
  }{\textheight}%
}{0.5ex}}%
\stackon[1pt]{#1}{\scalebox{-1}{\tmpbox}}%
}

\theoremstyle{plain} \newtheorem{lemma}{\textbf{Lemma}}
\newtheorem{proposition}{\textbf{Proposition}}

\newtheorem{theorem}{\textbf{Theorem}}
\newtheorem{assumption}{Assumption}

\theoremstyle{remark}\newtheorem{remark}{\textbf{Remark}}

\title{Exploration via Exploitation: The Blessing of Reward Diversity \\ in Personalized Federated RL}

 \author{
 	Jiin Woo\thanks{Work done while at Carnegie Mellon University. Correspondence to Jiin Woo (\texttt{jiinw@andrew.cmu.edu}).} \\
	CMU
 	\and
 	Gauri Joshi\thanks{Department of Electrical and Computer Engineering, Carnegie Mellon University.} \\
	CMU 
	\and
 	Yuejie Chi\thanks{Department of Statistics and Data Science, Yale University.} \\
	Yale
}

\date{}

\begin{document}

\maketitle

\begin{abstract}
While federated reinforcement learning (RL) has traditionally guaranteed high efficiency in homogeneous environments, collaboration is often hindered when agents possess distinct goals. In this paper, we flip this paradigm, framing reward heterogeneity not as a hurdle to consensus, but as a structural asset that catalyzes exploration. We first propose Personalized Federated Upper Confidence Bound Value Iteration (\pfucbvi), which achieves optimal linear speedup by decoupling shared dynamics from personalized objectives. To eliminate the risks of explicit exploration, we then introduce Personalized Federated Exploration-Free Value Iteration (\pfefvi), a purely greedy algorithm that leverages reward diversity to ensure state-action coverage. We prove that \pfefvi attains logarithmic regret without explicit exploration bonuses under sufficient reward diversity. Our results reveal that agent disagreement shifts the exploration burden from temporal complexity to spatial diversity, enabling efficient collective learning without explicit exploration.
\end{abstract}

\smallskip
\ifdefined\hidekeywords\else
\noindent\textbf{Keywords:} federated RL, personalized RL, online RL, exploration
\fi

\section{Introduction}
Reinforcement Learning (RL) has achieved remarkable success across various domains, yet its sample complexity remains a fundamental bottleneck in real-world applications \citep{chen2026statistical}. Federated Reinforcement Learning (FedRL) has emerged as a promising paradigm to alleviate this issue, enabling multiple agents to collaboratively learn policies by sharing experiences without centralizing raw trajectories \citep{khodadadian22fedrlspeedup,woo2023blessing,zheng2024federated}.

However, most FedRL literature assumes homogeneous environments where all agents share identical dynamics and rewards \citep{khodadadian22fedrlspeedup,woo2025blessing,zheng2024federated,zheng2025federated,salgia2024sample,zhang2025gap}, while in practice agents often pursue distinct, personalized goals. For instance, in a fleet of autonomous vehicles navigating the same city, each vehicle has a different destination, and recommendation systems on a common platform serve users with vastly different interests. Existing approaches for heterogeneous FedRL \citep{jin2022fedrlhete,wang2024convergence,mangold2025convergence,labbi2024federated,zhang2026personal,xiong2025linear,wang2026personalized} largely treat reward diversity as an obstacle to be tolerated, averaged out, or mitigated through restrictive structural assumptions, and consequently struggle to deliver personalized learning under arbitrary heterogeneity with provable guarantees. This paper takes the opposite view: the diversity of rewards is not a hurdle, but a structural resource that, when properly harnessed, can both accelerate collaborative learning and reshape the exploration-exploitation trade-off itself.

To put this view on rigorous footing, we consider a federated setting where each agent interacts with an episodic Markov Decision Process (MDP) with shared dynamics but diverse rewards. We first propose Personalized Federated Upper Confidence Bound Value Iteration (\pfucbvi), which decouples the shared dynamics from the personalized rewards: the central server globally aggregates empirical transitions to estimate dynamics, while each agent maintains strictly local UCB bonuses and Bellman backups tailored to its own reward. We prove that \pfucbvi achieves an optimal linear speedup in average per-agent regret with respect to the number of agents, fully agnostic to the level of reward heterogeneity. In other words, agents can pursue arbitrarily different goals without any heterogeneity penalty in the regret, and reward heterogeneity is no longer a barrier to collaboration.

However, in domains such as robotics or clinical trials, exploratory actions can be risky, unethical, or expensive. This concern motivates our central insight: in a multi-agent system, the heterogeneity of goals is itself a structural catalyst for exploration. If a population of agents possesses sufficiently diverse rewards, their collective greedy executions naturally span the state-action space, lifting the exploration burden from any individual agent and placing it on the diversity of the population as a whole. Building on this insight, we propose Personalized Federated Exploration-Free Value Iteration (\pfefvi), a purely greedy algorithm with neither UCB bonuses nor optimistic initialization. To formalize the condition under which collective coverage emerges, we introduce the {\em Reward Diversity} assumption, which posits that the agents' greedy policies jointly visit every state-action pair with non-trivial probability, even under imperfect transition kernel estimates. Under this assumption, \pfefvi attains an instance-dependent logarithmic regret, with the dependency on the episode count shrinking with the size and diversity of the ensemble.

It is worth noting that \pfefvi is not a free lunch but a conditional blessing: when rewards lack diversity, the ensemble's greedy executions can collapse. The two algorithms are therefore complementary: \pfucbvi serves as a worst-case safety net paying an explicit-exploration burn-in regardless of reward structure, while \pfefvi delivers the blessing of reward diversity precisely when the collective agent ensemble is rich enough. Our experiments confirm a clean phase transition consistent with this picture. We now summarize our main contributions below. Throughout, $M$ denotes the number of agents, $S$ the size of the state space, $A$ the size of the action space, $H$ the horizon, $K$ the number of episodes, $\lambda_{\mathsf{min}}$ the reward diversity parameter, and $\Delta_{\mathsf{min}}^m$ the minimum suboptimality gap of agent $m$, $1\leq m \leq M$.
\begin{itemize}
    \item \textbf{Personalized federated UCB achieving linear speedup:} We propose \pfucbvi, establishing that sharing transition dynamics while maintaining personalized UCB bonuses yields a near-optimal average regret $\tilde{\mathcal{O}}(\sqrt{SAKH^3/M})$ per agent. This proves that linear speedup can be achieved without any heterogeneity penalty in the bound, regardless of the severity of reward heterogeneity.
    \item \textbf{Diversity-induced global coverage:} We introduce the \textit{Reward Diversity} assumption (Assumption~\ref{assumption:expfree_global_coverage}), which posits that whenever agents act greedily under any estimated transition dynamics, the ensemble jointly visits every state-action pair with probability at least $\lambda_{\mathsf{min}} > 0$. Unlike standard coverage assumptions that rely on environment stochasticity or active exploration, ours is driven entirely by the diversity of agents' rewards. We identify two complementary structural regimes under which the assumption is provably feasible and support it empirically in random MDPs with diverse rewards across agents.
    
    \item \textbf{Exploration-free federated RL with reward diversity:} We propose \pfefvi, the first exploration-free algorithm for federated RL. We prove that under sufficient reward diversity, purely greedy agents achieve an instance-dependent regret bound of $\tilde{\mathcal{O}}\left(\frac{H^5 \log K}{M \lambda_{\mathsf{min}}} \sum_{m=1}^M \frac{1}{\Delta_{\mathsf{min}}^m}\right)$, where $\Delta_{\mathsf{min}}^m$ is agent $m$'s minimum suboptimality gap. The $\log K$ dependency is scaled by $1/(M \lambda_{\mathsf{min}})$, which shrinks as the ensemble grows and diversifies, signifying that the temporal exploration penalty is effectively absorbed by the diversity of the ensemble, fundamentally reshaping the exploration-exploitation trade-off.

\end{itemize}

Together, these results recast reward diversity from a hurdle to be tolerated into a structural resource for collective intelligence, turning divergent goals into a blessing for exploration-free, efficient federated decision making.

\section{Related Works}
\label{sec:relatedworks-full}

To situate our work within the broader context of federated RL, we review three critical research thrusts: homogeneous federated RL, heterogeneous federated RL, and exploration-free RL.

\paragraph{Federated RL with homogeneous environments.}
A large body of work on federated RL assumes homogeneous environments where all agents share the same MDP. For sample complexity, \citet{khodadadian22fedrlspeedup,woo2025blessing} establish linear speedup under generative or Markovian sampling. \citet{woo2024federated} extend this line to the offline setting under single-policy coverage, and \citet{jiao2026sample} extended to the average-reward setting. For online regret minimization in episodic tabular MDPs, \citet{zheng2024federated} propose a federated Q-learning algorithm with Hoeffding and Bernstein style UCB bonuses, achieving linear regret speedup with low communication cost, which \citet{zheng2025federated} subsequently improve via reference-advantage decomposition to attain near-optimal regret. \citet{salgia2024sample} characterize the fundamental sample-communication trade-off, while \citet{zhang2025gap} establish the gap-dependent regret bounds for online federated Q-learning. These results form the natural baseline, but
they do not address agents with personalized rewards.

\paragraph{Federated RL with heterogeneous environments.}
While the majority of federated RL literature assumes homogeneous environments, recent works have begun to explore the more realistic setting where agents possess different transition dynamics and reward functions. Existing literature in this area primarily focuses on two directions:
\begin{enumerate}

  \item \textbf{Consensus-based methods} \citep{wang2024momentum, zhang2024finite, labbi2024federated} learn a single global policy optimizing the average value across environments. Early baselines along this line include \citet{jin2022fedrlhete}, who introduce QAvg and PAvg under transition heterogeneity. \citet{wang2024convergence} subsequently provide a fine-grained analysis of federated Q-learning showing that local updates introduce a non-vanishing heterogeneity bias, and \citet{mangold2025convergence} prove that FedSARSA does not converge to the optimum of a virtual averaged environment but to an environment-coupled fixed point with a similar bias. From a stochastic approximation perspective, \citet{zhu2025achieving} achieve linear speedup under Markovian sampling without projection. On the policy gradient side, \citet{yang2024federated} develop federated natural policy gradient and actor critic methods under decentralized topology and reward heterogeneity, \citet{labbi2025global} reveal that under heterogeneity, optimal federated policies can be non-deterministic or even time-varying, and establish global convergence under local \L{}ojasiewicz conditions. Beyond cooperative settings, \citet{chen2023byzantine} consider Byzantine-robust distributed RL where a fraction of agents may report adversarial information. Despite their progress, all consensus-based methods inherently sacrifice individual utility for global agreement: as the average value can be far from any agent's optimum, the resulting policy can be sub-optimal for individual agents, and the benefits of collaboration shrink, and may turn into negative transfer, as heterogeneity grows.

\item \textbf{Personalized methods} instead allow agents to learn individualized policies. \citet{zhang2026personal} prove linear speedup via affinity-based variance reduction under similarity conditions, while \citet{xiong2025linear} and \citet{wang2026personalized} achieve linear speedup through shared low-dimensional representations for federated TD and average-reward TD learning, respectively. However, these methods either rely on similarity, low-rank, or shared-representation assumptions across agents, or assume a fixed behavior policy that already ensures sufficient coverage, and most analyses focus on policy evaluation rather than online regret minimization.

\end{enumerate}

Despite these advances, most personalized algorithms assume a degree of similarity between agents or rely on a fixed behavior policy that ensures sufficient coverage of the state-action space. Our work removes both restrictions: \pfucbvi is fully agnostic to the level of reward heterogeneity, and \pfefvi sources coverage from reward diversity rather than from a prescribed behavior policy.

\paragraph{Exploration-free learning and covariate diversity.}
In practice, many RL applications avoid explicit exploration in favor of a greedy approach, particularly in domains such as robotics or clinical trials where explorative actions may be risky, unethical, or prohibitively expensive. Beyond safety concerns, explorative solutions are often harder to implement, and their outcomes are less predictable or interpretable. Consequently, common sense often suggests a greedy strategy in real-world scenarios, even when standard theory dictates the necessity of exploration. Foundational studies by \citet{bastani2021expfree} and \citet{kannan2018smoothed} have investigated this phenomenon in linear contextual bandits. They established that under a sufficient diversity of contexts, known as the covariate-diversity assumption, a purely greedy selection can achieve sublinear regret without any explicit exploration mechanisms. Recently, \citet{luca2025explorationfree} further characterized the covariate-diversity condition for RL. However, their results still require strong environment stochasticity to ensure state-action coverage, which remains a stringent requirement for a single agent. To our knowledge,
no prior work simultaneously addresses (i) the federated heterogeneous
setting, (ii) online regret minimization, and (iii) exploration-free
learning; \pfefvi fills this gap by sourcing coverage from cross-agent
reward diversity rather than from environment stochasticity.

\section{Problem formulation}
\label{sec:problem}

We consider $M$ agents operating in $M$ finite-horizon MDPs
$\{\mathcal{M}^m\}_{m=1}^M$ with shared dynamics and diverse rewards. Each agent $m$ is characterized by a local MDP
\[
\mathcal{M}^m =
\big(\mathcal{S}, \mathcal{A}, H, \{P_h\}_{h=1}^H,
\{r_h^m\}_{h=1}^H \big).
\]
Here, $\mathcal{S}$ is a finite state space, $\mathcal{A}$ is a finite action
space, and $H$ is the episode horizon. The transition kernel
$P_h(\cdot \mid s,a) \in \Delta(\mathcal{S})$ specifies the distribution of the
next state at step $h$ after taking action $a \in \mathcal{A}$ in state
$s \in \mathcal{S}$. The agents share the same transition kernels
$\{P_h\}_{h=1}^H$, but each agent has its own
personalized reward function $r_h^m : \mathcal{S} \times \mathcal{A} \to [0,1]$. In this work, we assume the reward functions are known and fixed at each agent. A policy $\pi = \{\pi_h\}_{h=1}^H$ is a sequence of mappings $\pi_h : \mathcal{S} \to \Delta(\mathcal{A})$ that specifies the action selection rule at step $h$ for each state $s$. The value function of policy $\pi$ for agent $m$ at step $h$ is defined as
\begin{align}
V_h^{m, \pi} (s) = \mathbb{E} \left[ \sum_{h'=h}^H r_{h'}^m (s_{h'}, a_{h'}) \mid s_h = s \right],
\end{align}
where the expectation is taken over the randomness in the trajectory generated by following policy $\pi$ starting from state $s$ at step $h$. The optimal value function for agent $m$ is defined as $V_h^{m, \star} (s) = \sup_{\pi} V_h^{m, \pi} (s)$, and the optimal policy $\pi^{m, \star}$ is any policy that achieves this optimal value function at every $(h,s)$.

\paragraph{Goal: regret minimization in online RL.}
In online RL, for $K$ episodes, agents interact with their local environments using behavior policies
updated based on the data collected previously. At episode $k$, each agent $m$
selects a policy $\pi_k^m$ based on the data collected from previous episodes
and executes $\pi_k^m$ independently of the other agents to generate a trajectory
$\{s_{k,h}^m, a_{k,h}^m, r_{k,h}^m\}_{h=1}^H$, where the initial state
$s_{k,1}^m$ is drawn from $\rho$.

Unlike prior works that sacrifice individual utility for global consensus \citep{labbi2024federated,zhang2024finite}, in our setting, the goal of each agent is to learn their personalized optimal policy $\pi^{m,\star}$. The overall system aims to minimize the total regret across all agents over $K$ episodes:
\begin{align}
\mathsf{Regret} (K) = \sum_{m=1}^M \sum_{k=1}^K \mathbb{E}_{s_1 \sim \rho} \left [ V_1^{m, \star} (s_1) - V_1^{m, \pi_k^{m} } (s_1)  \right],
\end{align}
where $\pi_k^m$ is the policy of agent $m$ at episode $k$.

\section{Personalized federated online RL with UCB exploration}
\label{sec:pfucbvi}

To solve the personalized regret minimization problem efficiently, we must leverage the shared structure of the environments without forcing a single global objective. Since the transition kernel $P_h$ is common across all agents, sharing state-action visitation data allows the ensemble to estimate the dynamics significantly faster than any single agent could alone. 

However, because each agent pursues a distinct reward function $r_h^m$, their respective value functions, and consequently, their uncertainty about future returns, differ significantly. A standard federated approach with a single global exploration bonus would be heavily suboptimal here; it might force an agent to explore regions of the state space that are entirely irrelevant to its personalized goals. To bridge this gap, we propose {Personalized Federated Upper Confidence Bound Value Iteration (\pfucbvi)}. 

\subsection{Algorithm description}
The core principle of \pfucbvi is to globally aggregate empirical transition counts to build an accurate shared dynamics model, while strictly maintaining \textit{personalized} UCB bonuses and local Bellman backups tailored to each agent's individual objectives. The algorithm proceeds as follows.

\paragraph{Initialization:} the server initializes all global counts $N_h(s,a)$ and $N_h(s,a,s')$ to zero, and each agent $m \in [M]$ initializes $V_h^m(s) = Q_h^m(s,a) = H-h+1$ for all $(s,a,h)$ with $V_{H+1}^m = 0$.

\paragraph{Federated learning loop:} at each episode $k = 1, \dots, K$, the algorithm performs three steps:
\begin{enumerate}
\item \emph{Optimistic execution.} Agent $m$ follows the greedy policy 
$$\pi_{k,h}^{m}(s) = \arg\max_{a} Q_{h}^{m}(s, a)$$ 
to collect transitions, with ties broken uniformly at random once at the start of each episode.
\item \emph{Global aggregation.} The server updates the global counts via 
$N_h(s, a) \leftarrow N_h(s, a) + \sum_m \mathbb{I}\{(s_{k, h}^m, a_{k, h}^m) = (s, a)\}$ and $N_h(s, a, s')$ analogously, then computes and broadcasts the empirical transition kernel 
$\widehat{P}_h(s' \mid s, a) = N_h(s, a, s') / \max\{1, N_h(s, a)\}$.
\item \emph{Local Bellman backup.} For $h = H, \dots, 1$, each agent $m$ first computes its personalized bonus
\begin{align} \label{eq:vi_bonus}
B_{h}^m(s, a) = \sqrt{\frac{4 \Var_{\hat{P}_{h,s,a}}(V_{h+1}^{m}) \iota}{\max\{N_{h}(s,a), 1\}}} + \frac{25 S H^2 \iota}{\max\{N_{h}(s,a), 1\}}, \quad \iota = \log(28M^2SAKH/\delta),
\end{align}
and then performs the value iteration update for $h=H, \dots, 1$:
\begin{align*}
Q_{h}^{m}(s, a) &= \min\Big\{H-h+1,\; r_{h}^m(s, a) + B_{h}^m(s, a) + \sum_{s' \in \mathcal{S}} \hat{P}_{h}(s' \mid s, a) V_{h+1}^{m}(s')\Big\}, \\
V_{h}^{m}(s) &= \max_{a \in \mathcal{A}} Q_{h}^{m}(s, a).
\end{align*}
\end{enumerate}

To our knowledge, \pfucbvi is the first algorithm that combines online UCB exploration with personalization in federated RL. Prior personalized federated RL works focus on policy evaluation~\citep{xiong2025linear, wang2026personalized} or assume a fixed behavior policy with sufficient coverage~\citep{zhang2026personal}, sidestepping the exploration--exploitation tradeoff entirely. The federated UCBVI algorithm \citep{labbi2024federated}, on the other hand, performs online exploration but only learns a consensus policy. \pfucbvi sits at the intersection: a personalized extension of UCBVI~\citep{azar2017minimax} for online settings where each agent performs a personalized exploration and learns its own optimal policy under arbitrary reward heterogeneity.

\subsection{Regret analysis of \pfucbvi}
We now present the main theoretical result for \pfucbvi, which establishes a regret bound that achieves optimal linear speedup with respect to the number of agents $M$ while allowing for arbitrary reward heterogeneity.

\begin{theorem}[Regret of personalized federated UCBVI] \label{thm:fed_ucbvi_independent_regret}
Consider the setting of Section~\ref{sec:problem} with $M$ agents, $S$ states, $A$ actions, and a planning horizon $H$. Let the agents perform the \pfucbvi algorithm with a confidence parameter $\delta \in (0, 1)$.
Then, with probability at least $1-\delta$, the total regret satisfies:
\begin{equation}
\mathsf{Regret} (K) \leq C \cdot \left( \sqrt{SAMKH^3 \iota^2} + S^2 A H^3 \iota^2 \,+\,MSAH^2 \right),
\end{equation}
where $C > 0$ is a universal constant and $\iota=\log(28M^2SAKH/\delta)$.
\end{theorem}

The proof of Theorem~\ref{thm:fed_ucbvi_independent_regret} is provided in Appendix~\ref{sec:fed_ucbvi_independent_regret_proof}. The key implications of this result are as follows.

\paragraph{Optimal linear speedup and comparison with baselines.}
The total regret $\widetilde{\mathcal{O}}(\sqrt{SAMKH^3})$ in Theorem~\ref{thm:fed_ucbvi_independent_regret} translates into an average per-agent regret of $\widetilde{\mathcal{O}}(\sqrt{SAKH^3/M})$, so the number of episodes needed to reach a target average regret shrinks by a factor of $M$, the optimal linear speedup over isolated learning.
Beyond the rate itself, \pfucbvi differs from the closest baseline Fed-UCBVI~\citep{labbi2024federated} in both formulation and the role of heterogeneity. 
Fed-UCBVI targets a fundamentally different objective, learning a single \emph{consensus} policy whose suboptimality with respect to each agent's personalized optimum grows with the level of reward heterogeneity. In contrast, our regret bound is fully agnostic to the level of reward heterogeneity: it holds no matter how different the agents' rewards are, and in the homogeneous setting its leading rate matches that of~\citet{labbi2024federated} while each agent learns its own optimal policy. This shows that \pfucbvi exploits the shared transition structure to accelerate personalized learning, without paying any heterogeneity penalty in the bound.

\section{Personalized federated online RL without exploration}
\label{sec:pfefvi}

While \pfucbvi navigates reward heterogeneity through explicit UCB-based exploration, such exploratory actions remain a significant burden in safety-critical or resource-constrained domains. In fact, the same mechanism that enables heterogeneity-agnostic collaboration suggests a deeper opportunity: if agents pursuing different objectives naturally visit different regions of the environment, their collective self-interest can eliminate the need for explicit exploration altogether. This shifts our perspective on heterogeneity from a condition to be tolerated to a potent resource that provides exploration ``for free'' across agents, motivating the exploration-free paradigm we develop next.

\subsection{Key assumptions}
To enable exploration-free learning in this heterogeneous setting, we introduce a key structural assumption. While single-agent RL typically requires strong environment stochasticity (e.g., covariate diversity) \citep{luca2025explorationfree}, we show that reward heterogeneity can itself induce sufficient coverage. Let $d_h^{\pi}(s, a)$ denote the marginal visitation probability of observing state-action pair $(s, a)$ at step $h$ under policy $\pi$. 

\begin{assumption}[Reward Diversity] \label{assumption:expfree_global_coverage}
    Let $\mathcal{P}$ be the set of all valid transition kernels. For any assumed transition kernel $\tilde{P} \in \mathcal{P}$ and each agent $m \in [M]$, let $\tilde{Q}_{h}^m$ denote the action-value function computed using its true private reward $r^m$ and the transition $\tilde{P}$. Let $\tilde{\pi}^m$ be the corresponding greedy policy induced by $\tilde{Q}_{h}^m$ with uniform tie-breaking.
    We assume there exists a strictly positive constant $\lambda_{\mathsf{min}} > 0$ such that, for any assumed transition kernel $\tilde{P} \in \mathcal{P}$ and each step $h \in [H]$:
    \begin{align}
        \min_{(s,a) \in \mathcal{S}_h \times \mathcal{A}} \frac{1}{M} \sum_{m=1}^M d_h^{\tilde{\pi}^m}(s, a) \ge \lambda_{\mathsf{min}},
    \end{align}
    where $\mathcal{S}_h$ is the set of states reachable at step $h$, and $d_h^{\tilde{\pi}^m}(s, a)$ is the marginal visitation probability of state-action pair $(s,a)$ at step $h$ when executing policy $\tilde{\pi}^m$ in the \textit{true} environment $P$.
\end{assumption}

This structural assumption posits that the diversity of true goals is potent enough to prevent the ensemble's policies from collapsing into a narrow subspace, regardless of early estimation errors in the dynamics.
Since the sum of visitation probabilities across all state-action pairs is bounded, the most ideal scenario of perfectly uniform coverage yields $\lambda_{\mathsf{min}} \le 1/(|\mathcal{S}_h|A)$. Consequently, $1/\lambda_{\mathsf{min}}$ absorbs the state-action space complexity.

\paragraph{Reward-diversity-induced coverage relaxation.} Assumption~\ref{assumption:expfree_global_coverage} significantly relaxes the stringent requirements of single-agent RL by relying on the macroscopic diversity of the agents' goals rather than environment stochasticity. Intuitively, if $\{r^m\}_{m=1}^M$ are sufficiently diverse, their collective greedy executions will naturally span the state-action space, achieving exploration {\em for free} collectively. The parameter $\lambda_{\mathsf{min}}$ captures the degree of this system-wide heterogeneity; a larger $\lambda_{\mathsf{min}}$ indicates broader collective coverage, facilitating faster convergence, whereas a smaller $\lambda_{\mathsf{min}}$ implies more concentrated trajectories over limited state-action space.

\subsection{The feasibility of the reward diversity-induced coverage}
\label{sec:feasibility}

At the heart of Assumption~\ref{assumption:expfree_global_coverage} lies 
a simple mechanism: when agents' value functions point in 
\textit{sufficiently different directions} across the state space, any 
perturbation in the dynamics estimate $\tilde{P}$ pushes their greedy 
policies apart rather than together. To see this concretely, consider 
any assumed transition kernel $\tilde{P} \in \mathcal{P}$ and the 
Q-value gap between two actions $a, a' \in \mathcal{A}$:
\begin{align}
    \tilde{Q}_h^m(s, a) - \tilde{Q}_h^m(s, a')
    = \underbrace{\big[r_h^m(s, a) - r_h^m(s, a')\big]}_{
        \textsf{action preference}} 
    + \underbrace{\big[\tilde{P}_h(\cdot \mid s, a) 
    - \tilde{P}_h(\cdot \mid s, a')\big]^\top \tilde{V}_{h+1}^m}_{
        \textsf{dynamics-sensitive term}}.
    \label{eq:qvalue_decomposition}
\end{align}
The first term reflects each agent's intrinsic preference between 
actions and is invariant to dynamics estimation errors. The second term 
depends on $\tilde{P}$ through the agent's value function 
$\tilde{V}_{h+1}^m$ and is the source of $\tilde{P}$-dependent policy 
variation. Crucially, this second term is shaped by the geometry of 
$\tilde{V}_{h+1}^m$ and reward heterogeneity controls this geometry 
\textit{differently across agents}. When agents' value functions are 
diverse, the same dynamics perturbation steers different agents toward 
different actions, automatically maintaining collective coverage 
regardless of estimation errors.

We illustrate two complementary regimes where Assumption~\ref{assumption:expfree_global_coverage} is naturally 
satisfied, both ensuring that agents collectively cover all 
state-action pairs despite perturbations in $\tilde{P}$. The two 
regimes are distinguished by which term dominates the Q-value gap 
in~\eqref{eq:qvalue_decomposition}, and accordingly by how 
collective coverage is preserved. In one regime, each agent's 
reward gap dominates the dynamics-sensitive term, so individual 
choices stay strong under $\tilde{P}$-perturbations and agents with different reward preferences keep selecting different actions. In the 
other regime, the dynamics-sensitive term can be large and individual 
choices do shift with $\tilde{P}$, but cross-agent anti-correlation 
forces agents to shift in opposite directions so that different 
actions are always covered between them. In both cases, no matter 
how $\tilde{P}$ moves, the ensemble continues to spread its visits 
across the state-action space.

\begin{figure}[t]
    \centering
    \includegraphics[width=0.95\textwidth]{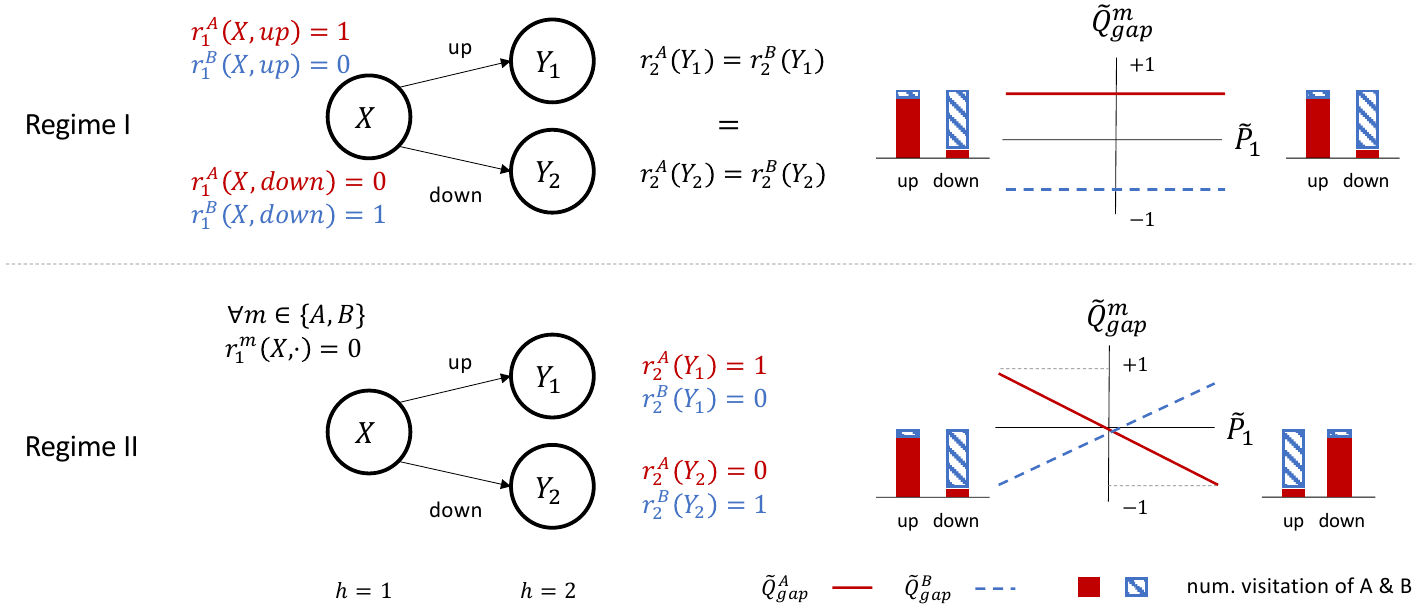}%
    \caption{Two regimes of reward heterogeneity that induce global 
coverage in Assumption~\ref{assumption:expfree_global_coverage}, 
illustrated on a common two-step MDP with 
$\mathcal{S} = \{x, Y_1, Y_2\}$, $\mathcal{A} = \{\textup{up}, 
\textup{down}\}$, $M = 2$ (agent A and agent B). 
In each row, the right panel plots each agent's step-1 \textit{Q-gap}
$\tilde{Q}^m_{gap} = \tilde{Q}_1^m(x, \textup{up}) - \tilde{Q}_1^m(x, \textup{down})$, 
whose sign determines the agent's greedy choice at $x$, as the 
assumed kernel $\tilde{P}_1$ varies.
\textbf{Top (Regime~I):} action-dependent state-independent rewards 
make $\tilde{V}_2^m$ take the same value at every state, so each 
agent's Q-gap is fixed regardless of $\tilde{P}_1$ 
(horizontal lines). Both agents persistently select different actions, 
and the cross-agent average visitation remains $1/2$ on each action at both extremes of $\tilde{P}_1$.
\textbf{Bottom (Regime~II):} state-dependent action-independent rewards 
anti-correlated across agents make $\tilde{V}_2^A + \tilde{V}_2^B$ 
take the same value at every state, so the two agents' Q-gaps are 
exact mirror images for any $\tilde{P}_1$ (crossing lines). 
Individual visitations swap between the two extremes (blue and red 
bars exchange positions), but the cross-agent average remains $1/2$ 
on each action throughout.}
    \label{fig:two-regimes}
\end{figure}

\paragraph{Regime~I: strong individual choices.}
Suppose each agent's reward gap $r_h^m(s, a) - r_h^m(s, a')$ is large 
relative to the spread of $\tilde{V}_{h+1}^m$ across states. Even when 
$\tilde{P}$ is perturbed and the dynamics-sensitive term shifts, the 
reward gap continues to dictate the sign of each agent's Q-gap (the 
left-hand side of \eqref{eq:qvalue_decomposition}, which determines 
the greedy choice between $a$ and $a'$), so each agent's greedy action 
does not flip. If different agents have reward gaps pointing toward 
different actions, they persistently select different actions across 
all $\tilde{P}$ and at every step $h$, and the ensemble jointly covers all reachable state-action pairs provided that each agent reaches every reachable state with positive probability. The defining feature is that no 
single agent's choice is sensitive to dynamics estimation errors. 
Figure~\ref{fig:two-regimes} (top) illustrates this with
action-dependent state-independent rewards on $\{x, Y_1, Y_2\}$: 
$\tilde{V}_{h+1}^m$ takes the same value at every state for any 
$\tilde{P}$, the dynamics-sensitive term vanishes identically, and 
each agent's Q-gap reduces to its reward gap.

\paragraph{Regime~II: coordinated opposition.}
Now suppose individual reward gaps are small while $\tilde{V}_{h+1}^m$ 
varies substantially across states. Each agent's greedy action 
\textit{does} respond to $\tilde{P}$, since the dynamics-sensitive 
term now controls the sign of the Q-gap. Coverage instead arises from 
cross-agent structure: when agents' value functions are 
\textit{anti-correlated}, in the sense that $\sum_m \tilde{V}_{h+1}^m$ 
is constant across states, the dynamics-sensitive terms of different 
agents are forced to sum to zero across agents so their Q-gaps become mirror 
images. The two agents may both flip their preferred action as 
$\tilde{P}$ changes, but they flip in lockstep opposition and always 
select different actions, again covering all reachable state-action 
pairs jointly. Figure~\ref{fig:two-regimes} (bottom) illustrates this
with state-dependent action-independent rewards on $\{x, Y_1, Y_2\}$, 
where agent~$A$ values $Y_1$ and agent~$B$ values $Y_2$: their Q-gaps are exact mirror images for any $\tilde{P}_1$, and they swap actions in perfect opposition as $\tilde{P}_1$ varies.

The two regimes show that the seemingly 
strong requirement of 
Assumption~\ref{assumption:expfree_global_coverage} is 
nonetheless attainable through plausible structures of reward 
heterogeneity. In Section~\ref{sec:experiments}, we find empirical support for the assumption in general random MDPs with diverse rewards.

\subsection{Algorithm description}

Based on the reward diversity assumption, we propose Personalized Federated Exploration-Free Value Iteration (\pfefvi). Unlike standard federated RL algorithms that require meticulous tuning of exploration bonuses, \pfefvi relies solely on the empirical average of the global transitions and pure greedy exploitations at the local level. The algorithm proceeds as follows.

\paragraph{Initialization:} the server initializes all global counts $N_h(s,a)$ and $N_h(s,a,s')$ to zero, and each agent $m$ initializes $V_h^m(s) = Q_h^m(s,a) = 0$ for all $(s,a,h)$ with $V_{H+1}^m(s) = 0$. Ties are broken uniformly at random.

\paragraph{Federated learning loop:} at each episode $k = 1, \dots, K$, the algorithm performs three steps:
\begin{enumerate}[leftmargin=*]
\item \emph{Local execution (pure exploitation).} Each agent $m \in [M]$ executes its current greedy policy $\pi_{k,h}^m(s) = \arg\max_a Q_h^m(s, a)$ to collect a local trajectory $\{s_{k,h}^m, a_{k,h}^m, r_h^m(s_{k,h}^m, a_{k,h}^m)\}_{h=1}^H$.
\item \emph{Global aggregation.} The server updates the global counts via $N_h(s, a) \leftarrow N_h(s, a) + \sum_m \mathbb{I}\{(s_{k,h}^m, a_{k,h}^m) = (s, a)\}$ and $N_h(s, a, s')$ analogously, then computes and broadcasts the empirical transition kernel $\widehat{P}_h(s' \mid s, a) = N_h(s, a, s') / N_h(s, a)$ if $N_h(s,a)\ge1$, and $\widehat{P}_h(s' \mid s, a)=1/S$ otherwise.
\item \emph{Local Bellman backup.} Each agent $m$ performs backward value iteration: for $h = H,  \ldots, 1$,
\begin{align*}
Q_h^m(s, a) &= r_h^m(s, a) + \sum_{s'} \widehat{P}_h(s' \mid s, a) V_{h+1}^m(s'), \\ 
V_h^m(s) &=  \max_a Q_h^m(s, a).
\end{align*}
\end{enumerate}

\subsection{Regret analysis of \pfefvi}

In this section, we present the theoretical guarantees of the \pfefvi algorithm. We show that under the reward diversity condition, a purely greedy approach without any explicit exploration or optimistic initialization achieves logarithmic cumulative regret.

\begin{theorem}[Instance-dependent regret bound for \pfefvi]\label{thm:main_regret}
    Suppose Assumption~\ref{assumption:expfree_global_coverage} holds with a global coverage parameter $\lambda_{\mathsf{min}} > 0$. When $M$ agents follow the \pfefvi algorithm with zero-initialization ($V_{1,h}^m = 0$), then with probability at least $1 - \delta$, the total system regret over $K$ episodes is bounded by:
    \begin{align}
        \mathsf{Regret} (K) \le C \cdot \left( \frac{H^5 \log(4 S A K M H / \delta)}{M \lambda_{\mathsf{min}}} \sum_{m=1}^M \frac{1}{\Delta_{\mathsf{min}}^m} + \frac{H \log(4 S A K M H / \delta)}{\lambda_{\mathsf{min}}} + MH \right),
    \end{align}
    where $C>0$ is a universal constant, and the instance-dependent gap $\Delta_{\mathsf{min}}^m$ for agent $m$ is defined as the minimum suboptimality gap across all state-action pairs, i.e., $\Delta_{\mathsf{min}}^m = \min_{(s,a,h): \Delta_h^m(s,a) > 0} \left( V_h^{m,\star}(s) - Q_h^{m,\star}(s, a) \right)$.
\end{theorem}

The proof of Theorem~\ref{thm:main_regret} is provided in Appendix~\ref{sec:proof-fedefvi-regret}. The key implications of this result are as follows.

\paragraph{Logarithmic regret without exploration.}
Theorem~\ref{thm:main_regret} establishes that \pfefvi attains an $\mathcal{O}(\log K)$ regret while every agent acts in a strictly greedy manner with respect to its own reward, without any UCB-style bonus, optimistic initialization, or random exploration. To our knowledge, this is the first gap-dependent, exploration-free regret analysis in the federated RL literature. The closest existing analyses leave a clear gap that our result fills. \citet{zhang2025gap} establish a gap-dependent regret of $\widetilde{\mathcal{O}}(H^6 SA \log T / \Delta_{\mathsf{min}})$ for federated online $Q$-learning, but only for \emph{homogeneous} MDPs and crucially through a UCB-based exploration mechanism. \pfefvi achieves a comparable $\widetilde{\mathcal{O}}(H^5 \log K / (\lambda_{\mathsf{min}} \Delta_{\mathsf{min}}))$ instance-dependent rate (here, we take $\Delta^m_{\mathsf{min}} = \Delta_{\mathsf{min}}$ for simplicity of comparison), without any exploration bonus and under arbitrary reward heterogeneity, with the state-action complexity $SA$ replaced by the effective coverage $1/\lambda_{\mathsf{min}}$ induced by reward diversity. A direct comparison with single-agent exploration-free RL is more delicate due to differences in assumptions: \citet{luca2025explorationfree} achieve a worst-case guarantee for linear MDPs with a single agent, but their result fundamentally relies on strong environment stochasticity to ensure state-action coverage, which is a stringent requirement for any single agent. \pfefvi removes this requirement entirely by sourcing coverage from cross-agent reward diversity rather than from the environment, so that even in deterministic or near-deterministic environments where their assumption is essentially vacuous, our $\log K$ guarantee continues to hold as long as $\lambda_{\mathsf{min}} > 0$.

\paragraph{From temporal exploration to spatial diversity.}
The coverage parameter $\lambda_{\mathsf{min}}$ enters the leading term of the regret bound only through the factor $1/(M \lambda_{\mathsf{min}})$, which can be interpreted as the inverse of an \emph{effective sample size} per episode: at every episode, the ensemble's diverse greedy choices already cover the state-action space at rate $M\lambda_{\mathsf{min}}$, so that the factor $SA$ in the $SA/\Delta_{\mathsf{min}}$ dependency of UCB-based methods is replaced by $1/\lambda_{\mathsf{min}}$. The two factors $M$ and $\lambda_{\mathsf{min}}$ are complementary rather than interchangeable: increasing $M$ alone, with all agents sharing the same reward, leaves $\lambda_{\mathsf{min}}$ at the coverage of a single greedy policy, which is typically zero, as Figure~\ref{fig:diversity-heatmap} illustrates empirically. The exploration cost is thus not eliminated but transferred from time to the population's reward diversity. The same $1/M$ prefactor also explains why the $\sum_m 1/\Delta^m_{\mathsf{min}}$ summation does not inflate the regret as one might fear: even an agent facing a hard task with very small $\Delta^m_{\mathsf{min}}$ inherits a federation-driven shrinkage of the global estimation error rather than relying on its own trajectories alone, and the per-agent cost of resolving a small gap decreases proportionally with $M$.

\paragraph{Complementary regimes with \pfucbvi.}
An order-wise regret comparison between \pfefvi and \pfucbvi is not directly meaningful: \pfucbvi enjoys an unconditional $\widetilde{\mathcal{O}}(\sqrt{SAMKH^3})$ guarantee that holds for any reward configuration, whereas \pfefvi's $\mathcal{O}(\log K)$ bound is conditioned on Assumption~\ref{assumption:expfree_global_coverage}: when $\lambda_{\mathsf{min}} = 0$, the ensemble's greedy executions can collapse onto a narrow subspace and \pfefvi may fail to identify the optimal policy at all. The two algorithms therefore occupy complementary regimes: \pfucbvi serves as a worst-case safety net, paying an explicit-exploration burn-in regardless of how rich the rewards are, while \pfefvi enjoys a free lunch precisely when the population's reward diversity is strong enough that Assumption~\ref{assumption:expfree_global_coverage} is satisfied with non-trivial $\lambda_{\mathsf{min}}$. In this regime, \pfefvi matches or outperforms \pfucbvi not by improving the regret but by avoiding the burn-in cost of its explicit UCB bonus, a phase transition empirically observed in Section~\ref{sec:performance-experiments}.

{\paragraph{Visitation-weighted regret bound.} The worst-pair factor $1/\lambda_{\mathsf{min}}$ in the leading term of Theorem~\ref{thm:main_regret} can be replaced by a visitation-weighted coefficient, as follows; the proof is given in Appendix~\ref{sec:proof-fedefvi-weighted}.
For each $(h,s,a)$, let
\begin{align}
\lambda_h(s,a):=\inf_{\tilde P\in\mathcal P}\ \frac1M\sum_{m=1}^M d_h^{\tilde\pi^m}(s,a),\label{eq:wt_dlow}
\end{align}
where $\tilde\pi^m$ is the uniformly tie-broken greedy policy of Assumption~\ref{assumption:expfree_global_coverage}. Assumption~\ref{assumption:expfree_global_coverage} is equivalent to $\min_{h,s,a}\lambda_h(s,a)\ge\lambda_{\mathsf{min}}$.
For a policy $\pi$, $\mexp^{\pi}[\,\cdot\mid s_h=s,a_h=a]$ is the expectation over the trajectory generated from $(s_h,a_h)=(s,a)$ in the true environment by $s_{h'+1}\sim P_{h'}(\cdot|s_{h'},a_{h'})$ and $a_{h'+1}\sim\pi_{h'+1}(\cdot|s_{h'+1})$; the expectation $\mexp^{\pi}[\,\cdot\mid s_h=s]$ additionally draws $a_h\sim\pi_h(\cdot|s)$.
For each agent $m$, define
\begin{align}
\Lambda^m:=\sup_{\tilde P\in\mathcal P}\Bigg\{\frac1H\sum_{h=1}^H\mexp_{(s,a)\sim d_h^{\tilde\pi^m}}\Big[\frac{1}{\lambda_h(s,a)}\Big]+\frac1{H^2}\sum_{h=1}^H\mexp_{s\sim d_h^{\tilde\pi^m}}\mexp^{\pi^{m,\star}}\Big[\sum_{h'=h}^H\frac{1}{\lambda_{h'}(s_{h'},a_{h'})}\,\Big|\,s_h=s\Big]\Bigg\},\label{eq:wt_Lambda}
\end{align}
where $d^{\pi}_h(s)=\sum_ad^{\pi}_h(s,a)$ is the state marginal. Since $1/\lambda_h(s,a)\le1/\lambda_{\mathsf{min}}$ for all pairs,
\begin{align}
\Lambda^m\le\frac1{\lambda_{\mathsf{min}}}\Big(1+\frac1{H^2}\sum_{h=1}^H(H-h+1)\Big)\le\frac{2}{\lambda_{\mathsf{min}}}.\label{eq:wt_Lambda_bound}
\end{align}
The first term of $\Lambda^m$ averages the inverse coverage along agent $m$'s own greedy trajectories. The second term averages it along the trajectories that $\pi^{m,\star}$ would follow from the states agent $m$ reaches. These are the pairs whose estimates determine whether a greedy action is recognized as suboptimal.

\begin{proposition}\label{prop:wt_regret}
Under the conditions of Theorem~\ref{thm:main_regret}, with probability at least $1-\delta$,
\begin{align}
\mathsf{Regret}(K)\le C\left[\log(1+K_0)\left(\frac{H^5\iota}{M}\sum_{m=1}^M\frac{\Lambda^m}{\Delta^m_{\mathsf{min}}}+\frac{H^4\sqrt S\,\iota}{\lambda_{\mathsf{min}}}\right)+\frac{H\iota}{\lambda_{\mathsf{min}}}+MH\right],\label{eq:wt_main}
\end{align}
where $C$ is a universal constant and $K_0=\max_mK_0^m$ with $K_0^m=\big\lceil C_0H^4\iota/(M\lambda_{\mathsf{min}}(\Delta^m_{\mathsf{min}})^2)\big\rceil$ for a sufficiently large universal constant $C_0$.
\end{proposition}

\begin{remark}
Compared with Theorem~\ref{thm:main_regret}, the worst-pair factor $1/\lambda_{\mathsf{min}}$ in the leading, gap-dependent term is replaced by $\Lambda^m\le2/\lambda_{\mathsf{min}}$, the inverse coverage averaged along agent $m$'s greedy trajectories and the optimal continuations from the states it reaches. This is an improvement when the worst-covered pairs are rarely on these trajectories, i.e., $\Lambda^m\ll1/(\lambda_{\mathsf{min}}\log(1+K_0))$, and the gaps are small enough for the leading term to dominate. The burn-in and gap-independent terms still scale with $1/\lambda_{\mathsf{min}}$, so Assumption~\ref{assumption:expfree_global_coverage} is still required.
\end{remark}
}

\section{Experiments}
\label{sec:experiments}

We complement the theoretical results with two empirical studies on a single unified family of synthetic MDPs: Section~\ref{sec:feasibility-experiments} provides empirical support for Assumption~\ref{assumption:expfree_global_coverage} under generic reward heterogeneity beyond the structural regimes of Section~\ref{sec:feasibility}, and Section~\ref{sec:performance-experiments} compares \pfucbvi{} and \pfefvi{} as $M$ varies, exhibiting a phase transition qualitatively consistent with our theory.

\subsection{Experimental setup}
\label{sec:setup}

We use synthetic tabular MDPs with $P_h(\cdot \mid s, a) \sim \text{Dirichlet}(\alpha \cdot \mathbf{1})$ and sparse heterogeneous rewards: each agent receives reward $1$ on an independently drawn random subset of $\lceil \kappa \cdot SAH \rceil$ cells $(h,s,a)$ and $0$ elsewhere, where small $\alpha$ and $\kappa$ make myopic behavior unlikely to reach the rewarding cells. The parameters and implementation details are given in Appendix~\ref{sec:exp-details}.

\subsection{Empirical feasibility under generic reward heterogeneity}
\label{sec:feasibility-experiments}

The two regimes in Section~\ref{sec:feasibility} establish that Assumption~\ref{assumption:expfree_global_coverage} is attainable under specific structural conditions. We now examine it empirically under \emph{generic} reward heterogeneity, where each agent's preferred reward cells are drawn independently and neither regime applies by construction. To probe robustness to dynamics estimation error, we perturb the true kernel as $\tilde{P} = (1-\epsilon) P + \epsilon U$ with $U_h(\cdot \mid s, a) \sim \text{Dirichlet}(\mathbf{1})$, and measure $\hat{\lambda}_{\min}(\tilde{P}) = \min_{(s,a,h)} \frac{1}{M} \sum_{m=1}^M d_h^{\tilde{\pi}^m}(s,a)$.

\begin{figure}[t]
\centering
\includegraphics[width=\ifdefined\perfsplit 0.8\else 0.9\fi\textwidth]{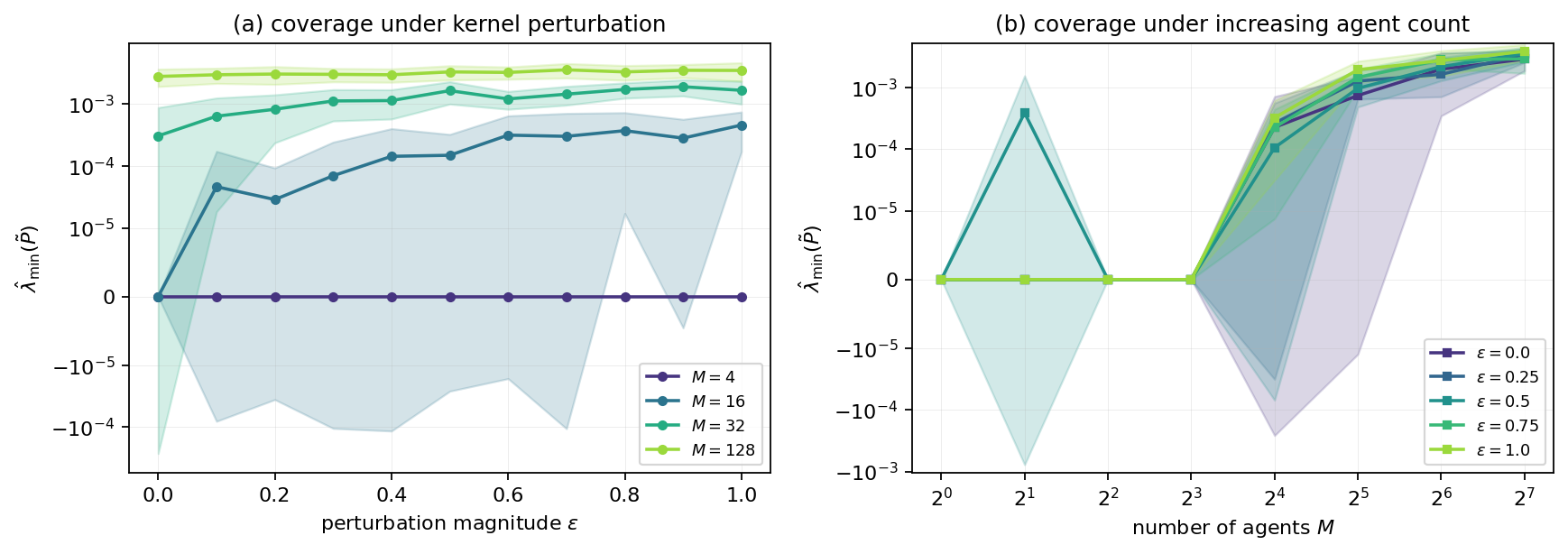}%
\caption{Empirical feasibility of Assumption~\ref{assumption:expfree_global_coverage} on random tabular MDPs. \textbf{(a)} $\hat{\lambda}_{\min}(\tilde{P})$ shows no downward trend in $\epsilon$. \textbf{(b)} $\hat{\lambda}_{\min}$ becomes positive from $M=16$ and grows with $M$, overlapping across $\epsilon$, indicating that reward heterogeneity drives coverage independently of $\tilde{P}$.}
\label{fig:feasibility}
\end{figure}

Figure~\ref{fig:feasibility} reports the result. The curves show no downward trend in $\epsilon$, indicating that collective coverage is not destroyed by dynamics estimation error, and from $M=16$ on $\hat{\lambda}_{\min}$ grows with $M$ with curves overlapping across $\epsilon$, indicating that coverage is driven by reward heterogeneity rather than by the kernel perturbation. To rule out that this simply reflects the number of policies executed, Figure~\ref{fig:diversity-heatmap} compares two reward setups at $M=64$: identical rewards yield identical policies and highly uneven visitation, whereas i.i.d.\ rewards spread visits nearly uniformly across $(s,h)$. The contrast isolates reward diversity, not ensemble size, as the source of coverage.

\ifdefined\heatmapinappendix\else\providecommand{\heatmapplacement}{t}%
\edef\heatmapfigbegin{\noexpand\begin{figure}[\heatmapplacement]}\heatmapfigbegin
\centering
\includegraphics[width=0.75\textwidth]{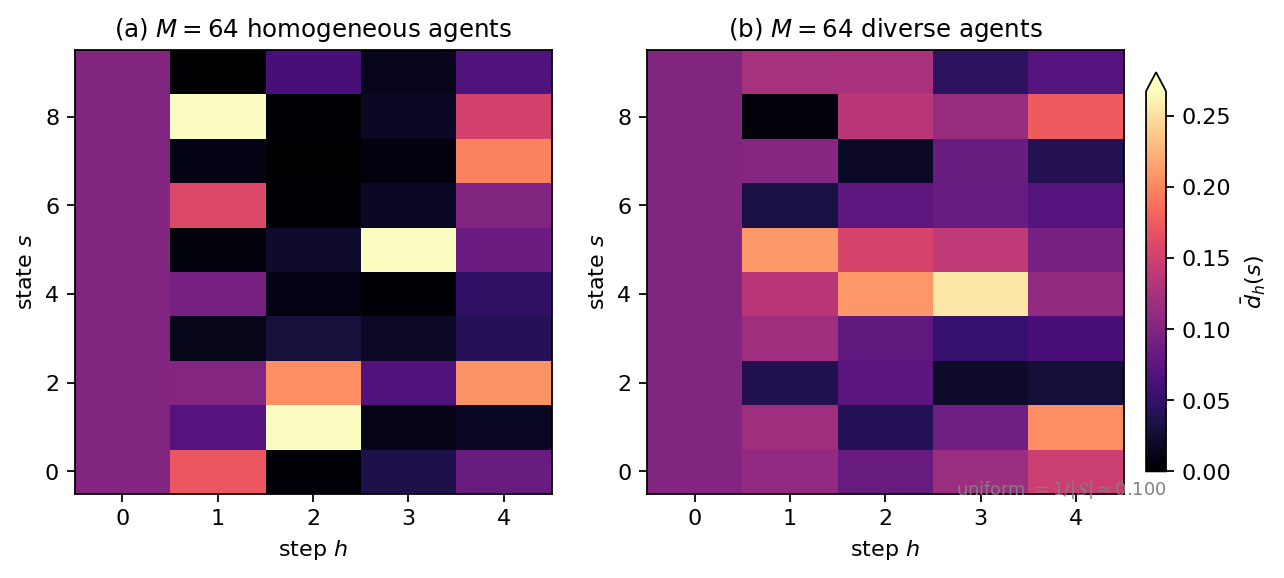}
\caption{Reward diversity, not ensemble size, drives collective coverage. State visitation $\bar d_h(s)$ on a single MDP with $\tilde P = P$. \textbf{(a)} $M=64$ agents with identical rewards: visitation is highly uneven (brightest cells exceed the color scale). \textbf{(b)} $M=64$ agents with i.i.d.\ rewards: visits spread nearly uniformly. The shared color scale is capped at the diverse panel's maximum.}
\label{fig:diversity-heatmap}
\end{figure}
\fi

\subsection{Performance of \pfucbvi{} and \pfefvi{}}
\label{sec:performance-experiments}

\begin{figure}[t]
\centering
\includegraphics[width=0.9\textwidth]{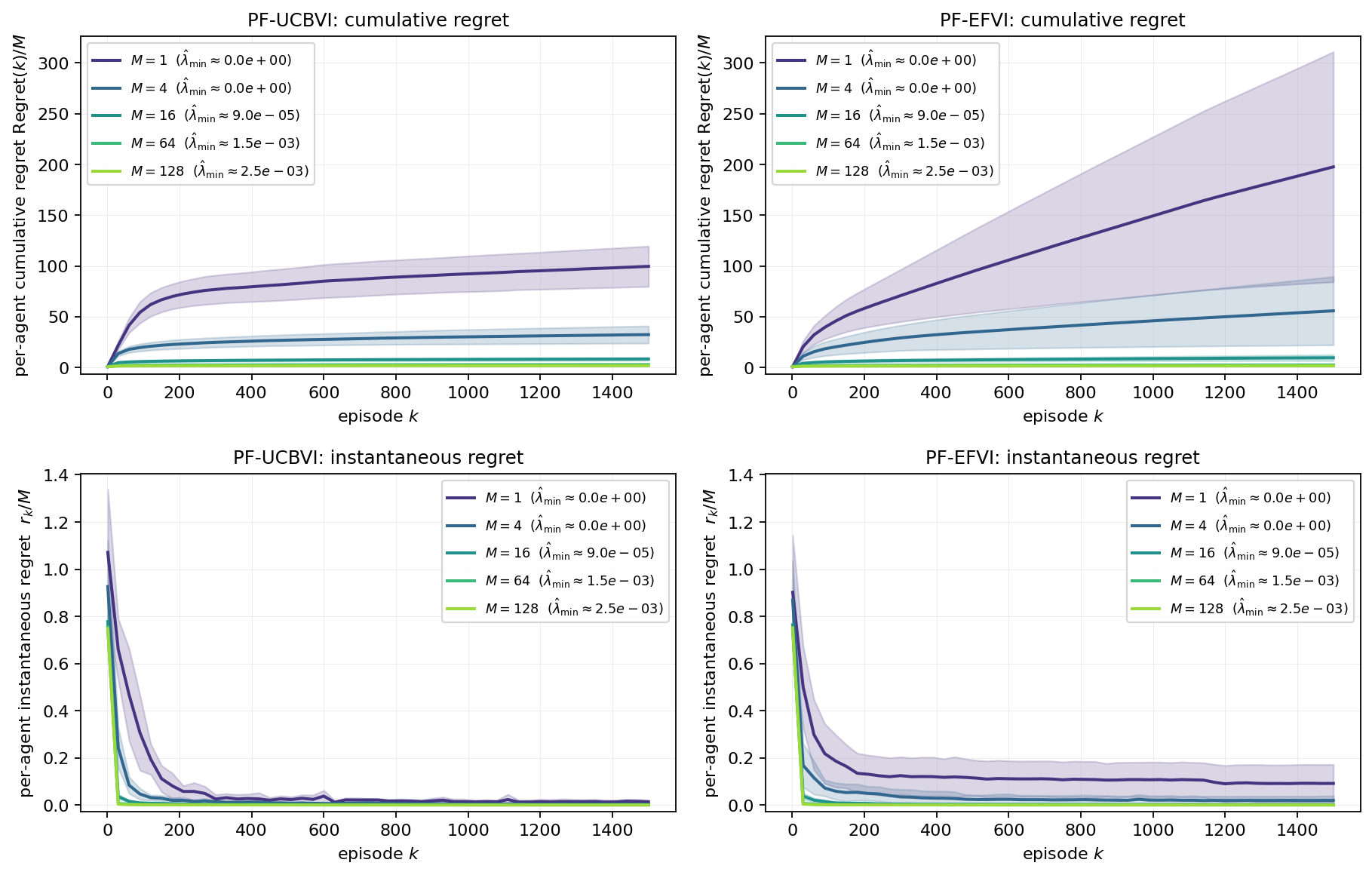}
\caption{Per-agent regret of \pfucbvi{} and \pfefvi{}. \textbf{Top:} cumulative regret $\textnormal{Regret}(k)/M$. \textbf{Bottom:} instantaneous regret. \pfucbvi{} converges sub-linearly at every $M$ with a visible burn-in. \pfefvi{} fails at small $M$ ($\widehat{\lambda}_{\min}=0$ for $M=1,4$) but matches or beats \pfucbvi{} once $M \geq 16$ by avoiding the explicit-exploration burn-in.}
\label{fig:regret}
\end{figure}

We compare the per-agent cumulative regret of \pfucbvi{} and \pfefvi{} on the unified MDP family of Section~\ref{sec:setup}, in which the myopic $r$-greedy policy is far from optimal. We sweep $M \in \{1, 4, 16, 64, 128\}$ over $K = 1500$ episodes and report results averaged over $8$ MDP draws\ifdefined\perfsplit\else{} with shaded bands showing one standard deviation\fi. \ifdefined\perfsplit Figure~\ref{fig:regret} displays the per-agent cumulative regret $\textnormal{Regret}(k)/M$.\else Figure~\ref{fig:regret} displays both per-agent cumulative regret $\textnormal{Regret}(k)/M$ (top) and per-agent instantaneous regret (bottom).\fi

The results support the theory along three axes. First, \ifdefined\perfsplit \pfucbvi{}'s cumulative regret flattens at every $M$\else \pfucbvi{}'s instantaneous regret decays toward zero at every $M$\fi, faster for larger $M$, qualitatively consistent with the linear speedup of Theorem~\ref{thm:fed_ucbvi_independent_regret}. Second, at $M = 1, 4$ the empirical $\widehat{\lambda}_{\min}$ is identically zero across all $8$ MDP draws: \pfefvi{}'s greedy agents cannot escape the narrow region induced by their own policies, \ifdefined\perfsplit its cumulative regret keeps growing linearly, reaching roughly twice that of \pfucbvi{} by $K=1500$ at $M=1$\else its instantaneous regret plateaus well above zero, and its cumulative regret reaches roughly twice that of \pfucbvi{} by $K=1500$ at $M=1$\fi, indicating that exploitation alone can fail when reward heterogeneity is insufficient. Third, once $M \geq 16$, $\widehat{\lambda}_{\min}$ becomes positive and grows with $M$, and \pfefvi{}'s per-agent regret matches \pfucbvi{} (within standard deviations at $M=16, 64$, strictly below at $M=128$). The advantage at large $M$ is precisely the burn-in cost \pfefvi{} avoids by relying on heterogeneous greedy behaviours instead of explicit bonuses, in line with the temporal-to-spatial trade-off of Theorem~\ref{thm:main_regret}. A comparison with Fed-UCBVI is in Appendix~\ref{sec:exp-details}.

\section{Discussion}
This work argues that reward heterogeneity is a structural advantage in FedRL. \pfucbvi{} demonstrates that decoupling shared dynamics from personalized rewards ensures optimal linear speedup with no heterogeneity penalty in the regret bound. Moving beyond linear speedup, \pfefvi{} reveals that spatial diversity can eliminate the need for risky, explicit exploration altogether. By transferring the exploration burden from time to the scale of the ensemble, we show that purely greedy agents can achieve sufficient exploration and efficient convergence to the optimal policy. This blessing, however, is a conditional one: when the population's rewards are not diverse enough, the ensemble's greedy executions can collapse and exploitation alone is no longer sufficient. \pfucbvi{} and \pfefvi{} are therefore complementary rather than substitutes, and learning when to rely on each, or how to interpolate between them, is a natural direction for future work. More broadly, our findings redefine heterogeneity as a primary resource for collective intelligence, turning divergent goals into a catalyst for exploration-free efficiency. Our work opens up many directions for future research, such as relaxing the reward diversity assumption and exploring the interplay between reward and dynamics heterogeneity. It will also be of great interest to extend our results to the setting with general function approximation, and to empirically validate the benefits of reward diversity in real-world applications of FedRL.

\section*{Acknowledgements}
The work of G. Joshi is supported in part by the grants NSF-CCF 2007834, CCF-2045694, CNS-2112471, and ONR N00014-23-1-2149.
The work of Y. Chi is supported in part by NSF under grants ECCS-2537189 and CNS-2148212, and in part by funds from federal agency and industry partners as specified in the Resilient \& Intelligent NextG Systems (RINGS) program.

\bibliographystyle{apalike} 
\bibliography{bibfileRL}

\begin{thebibliography}{}

\bibitem[Azar et~al., 2017]{azar2017minimax}
Azar, M.~G., Osband, I., and Munos, R. (2017).
\newblock Minimax regret bounds for reinforcement learning.
\newblock In {\em International Conference on Machine Learning}, pages
  263--272. JMLR. org.

\bibitem[Bastani et~al., 2021]{bastani2021expfree}
Bastani, H., Bayati, M., and Khosravi, K. (2021).
\newblock Mostly exploration-free algorithms for contextual bandits.
\newblock {\em Management Science}, 67(3):1329--1349.

\bibitem[Chen et~al., 2026]{chen2026statistical}
Chen, Y., Chi, Y., Fan, J., Li, G., Wei, Y., and Yan, Y. (2026).
\newblock Statistical foundations of reinforcement learning: A non-asymptotic
  perspective.
\newblock {\em arXiv preprint arXiv:2610.41903}.

\bibitem[Chen et~al., 2023]{chen2023byzantine}
Chen, Y., Zhang, X., Zhang, K., Wang, M., and Zhu, X. (2023).
\newblock Byzantine-robust online and offline distributed reinforcement
  learning.
\newblock In {\em International Conference on Artificial Intelligence and
  Statistics}, pages 3230--3269. PMLR.

\bibitem[Civitavecchia and Papini, 2025]{luca2025explorationfree}
Civitavecchia, L. and Papini, M. (2025).
\newblock Exploration-free reinforcement learning with linear function
  approximation.
\newblock In {\em Reinforcement Learning Conference}.

\bibitem[Dann et~al., 2017]{dann2017unifying}
Dann, C., Lattimore, T., and Brunskill, E. (2017).
\newblock Unifying {PAC} and regret: Uniform {PAC} bounds for episodic
  reinforcement learning.
\newblock {\em arXiv preprint arXiv:1703.07710}.

\bibitem[Jiao et~al., 2026]{jiao2026sample}
Jiao, Y., Woo, J., Li, G., Joshi, G., and Chi, Y. (2026).
\newblock Sample complexity of average-reward {Q}-learning: From single-agent
  to federated reinforcement learning.
\newblock {\em arXiv preprint arXiv:2601.13642}.

\bibitem[Jin et~al., 2022]{jin2022fedrlhete}
Jin, H., Peng, Y., Yang, W., Wang, S., and Zhang, Z. (2022).
\newblock Federated reinforcement learning with environment heterogeneity.
\newblock In {\em International Conference on Artificial Intelligence and
  Statistics}, pages 18--37.

\bibitem[Kannan et~al., 2018]{kannan2018smoothed}
Kannan, S., Morgenstern, J., Roth, A., Waggoner, B., and Wu, Z.~S. (2018).
\newblock A smoothed analysis of the greedy algorithm for the linear contextual
  bandit problem.
\newblock In {\em Proceedings of the 32nd International Conference on Neural
  Information Processing Systems}.

\bibitem[Khodadadian et~al., 2022]{khodadadian22fedrlspeedup}
Khodadadian, S., Sharma, P., Joshi, G., and Maguluri, S.~T. (2022).
\newblock Federated reinforcement learning: Linear speedup under {M}arkovian
  sampling.
\newblock In {\em International Conference on Machine Learning}, pages
  10997--11057.

\bibitem[Labbi et~al., 2025a]{labbi2025global}
Labbi, S., Mangold, P., Tiapkin, D., and Moulines, E. (2025a).
\newblock On global convergence rates for federated policy gradient under
  heterogeneous environment.
\newblock {\em arXiv preprint arXiv:2505.23459}.

\bibitem[Labbi et~al., 2025b]{labbi2024federated}
Labbi, S., Tiapkin, D., Mancini, L., Mangold, P., and Moulines, {\'E}. (2025b).
\newblock Federated {UCBVI}: Communication-efficient federated regret
  minimization with heterogeneous agents.
\newblock In {\em Proceedings of The 28th International Conference on
  Artificial Intelligence and Statistics}, volume 258 of {\em Proceedings of
  Machine Learning Research}, pages 1315--1323. PMLR.

\bibitem[Mangold et~al., 2025]{mangold2025convergence}
Mangold, P., Berthier, E., and Moulines, E. (2025).
\newblock Convergence guarantees for federated {SARSA} with local training and
  heterogeneous agents.
\newblock {\em arXiv preprint arXiv:2512.17688}.

\bibitem[Maurer and Pontil, 2009]{maurer2009empirical}
Maurer, A. and Pontil, M. (2009).
\newblock Empirical bernstein bounds and sample variance penalization.
\newblock In {\em Proc. Computational Learning Theory Conference}.

\bibitem[Salgia and Chi, 2024]{salgia2024sample}
Salgia, S. and Chi, Y. (2024).
\newblock The sample-communication complexity trade-off in federated
  {Q}-learning.
\newblock In {\em The Thirty-eighth Annual Conference on Neural Information
  Processing Systems}.

\bibitem[Wang et~al., 2024a]{wang2024momentum}
Wang, H., He, S., Zhang, Z., Miao, F.~M., and Anderson, J. (2024a).
\newblock Momentum for the win: collaborative federated reinforcement learning
  across heterogeneous environments.
\newblock In {\em Proceedings of the 41st International Conference on Machine
  Learning}, ICML'24. JMLR.org.

\bibitem[Wang et~al., 2026]{wang2026personalized}
Wang, L.~M., Yang, P., and Su, L. (2026).
\newblock Personalized multi-agent average reward {TD}-learning via joint
  linear approximation.
\newblock {\em arXiv preprint arXiv:2603.02426}.

\bibitem[Wang et~al., 2024b]{wang2024convergence}
Wang, M., Yang, P., and Su, L. (2024b).
\newblock On the convergence rates of federated {Q}-learning across
  heterogeneous environments.
\newblock {\em arXiv preprint arXiv:2409.03897}.

\bibitem[Weissman et~al., 2003]{weissman2003inequalities}
Weissman, T., Ordentlich, E., Seroussi, G., Verd{\'u}, S., and Weinberger,
  M.~J. (2003).
\newblock Inequalities for the {$L_1$} deviation of the empirical distribution.
\newblock Technical Report HPL-2003-97, Hewlett-Packard Labs.

\bibitem[Woo et~al., 2023]{woo2023blessing}
Woo, J., Joshi, G., and Chi, Y. (2023).
\newblock The blessing of heterogeneity in federated {Q}-learning: Linear
  speedup and beyond.
\newblock In {\em Proceedings of the 40th International Conference on Machine
  Learning}, volume 202 of {\em Proceedings of Machine Learning Research},
  pages 37157--37216. PMLR.

\bibitem[Woo et~al., 2025]{woo2025blessing}
Woo, J., Joshi, G., and Chi, Y. (2025).
\newblock The blessing of heterogeneity in federated {Q}-learning: Linear
  speedup and beyond.
\newblock {\em Journal of Machine Learning Research}, 26(26):1--85.

\bibitem[Woo et~al., 2024]{woo2024federated}
Woo, J., Shi, L., Joshi, G., and Chi, Y. (2024).
\newblock Federated offline reinforcement learning: Collaborative single-policy
  coverage suffices.
\newblock In {\em Proceedings of the 41st International Conference on Machine
  Learning}, volume 235, pages 53165--53201. PMLR.

\bibitem[Xiong et~al., 2025]{xiong2025linear}
Xiong, G., Wang, S., Jiang, D., and Li, J. (2025).
\newblock On the linear speedup of personalized federated reinforcement
  learning with shared representations.
\newblock In {\em The Thirteenth International Conference on Learning
  Representations}.

\bibitem[Yang et~al., 2024]{yang2024federated}
Yang, T., Cen, S., Wei, Y., Chen, Y., and Chi, Y. (2024).
\newblock Federated natural policy gradient and actor critic methods for
  multi-task reinforcement learning.
\newblock In {\em The Thirty-eighth Annual Conference on Neural Information
  Processing Systems}.

\bibitem[Zhang and Azizan, 2026]{zhang2026personal}
Zhang, C. and Azizan, N. (2026).
\newblock Personalized collaborative learning with affinity-based variance
  reduction.
\newblock In {\em The Fourteenth International Conference on Learning
  Representations}.

\bibitem[Zhang et~al., 2024]{zhang2024finite}
Zhang, C., Wang, H., Mitra, A., and Anderson, J. (2024).
\newblock Finite-time analysis of on-policy heterogeneous federated
  reinforcement learning.
\newblock In {\em The Twelfth International Conference on Learning
  Representations}.

\bibitem[Zhang et~al., 2025]{zhang2025gap}
Zhang, H., Zheng, Z., and Xue, L. (2025).
\newblock Gap-dependent bounds for federated $q$-learning.
\newblock In {\em Proceedings of the 42nd International Conference on Machine
  Learning}.

\bibitem[Zheng et~al., 2024]{zheng2024federated}
Zheng, Z., Gao, F., Xue, L., and Yang, J. (2024).
\newblock Federated {Q}-learning: Linear regret speedup with low communication
  cost.
\newblock In {\em The Twelfth International Conference on Learning
  Representations}.

\bibitem[Zheng et~al., 2025]{zheng2025federated}
Zheng, Z., Zhang, H., and Xue, L. (2025).
\newblock Federated q-learning with reference-advantage decomposition: Almost
  optimal regret and logarithmic communication cost.
\newblock In {\em ICLR}.

\bibitem[Zhu et~al., 2025]{zhu2025achieving}
Zhu, F., Mitra, A., and Heath, R.~W. (2025).
\newblock Achieving tighter finite-time rates for heterogeneous federated
  stochastic approximation under markovian sampling.
\newblock {\em arXiv preprint arXiv:2504.11645}.

\end{thebibliography}

\appendix

\section{Notation}
Throughout the analysis, $X_{k,h}$ for $X \in \{N, \widehat P, B^m, Q^m, V^m\}$ denotes the value of $X$ at the \emph{beginning of episode $k$} (i.e., before episode $k$'s 
execution step). All these values are computed using data from episodes 
$1, \ldots, k-1$.
In particular:
\begin{itemize}
\item $X_{1, h}$ refers to the initialization (no data collected yet);
\item $N_{k, h}(s, a) = \sum_{m=1}^M \sum_{i=1}^{k-1} \mathbb{I}\{(s_{i, h}^m, a_{i, h}^m) = (s, a)\}$;
\item $\widehat P_{k, h}(s'|s, a) = N_{k, h}(s, a, s') / \max\{1, N_{k, h}(s, a)\}$. For \pfefvi, $\widehat P_{k,h}(\cdot|s,a)$ is instead the uniform distribution when $N_{k,h}(s,a)=0$
\item $\widehat P_{k, h, s, a} = \widehat P_{k, h}(\cdot|s, a)$ and $P_{h, s, a} = P_h(\cdot|s, a)$.
\item $\iota=\log(28M^2SAKH/\delta)$.

\end{itemize}
The policy executed in episode $k$ is $\pi_{k, h}^m(s) = \arg\max_a Q_{k, h}^m(s, a)$. For \pfucbvi, the tie-breaking draw of episode $k$ is made before the episode and is included in the filtration $\mathcal F_\tau$ of Step~4 from the start of episode $k$, so $\pi^m_k$ is a deterministic policy given the history. For \pfefvi, ties are broken uniformly at random, and $\pi^m_{k,h}(\cdot|s)$ denotes the resulting uniform distribution over $\argmax_aQ^m_{k,h}(s,a)$.

\section{Proof of Theorem~\ref{thm:fed_ucbvi_independent_regret}}
\label{sec:fed_ucbvi_independent_regret_proof}
To analyze the regret of \pfucbvi, we bound the total expected regret as follows:
$$\mathsf{Regret}(K) = \sum_{m=1}^M\sum_{k=1}^K\mexp_{s_1\sim\rho} \left[V_1^{m, \star}(s_1) - V_1^{m, \pi_k^m}(s_1) \right],$$
where, for each agent $m$ and episode $k$, we write $\Delta^{m}_{k,h}(s) = V_{k,h}^m(s) - V_h^{m, \star}(s)$ and $\Delta^{m,\pi_k^m}_{k,h}(s) = V_{k,h}^m(s) - V_h^{m, \pi_k^m}(s)$.
We first prove the optimism of the value function, i.e., $\Delta_{k,h}^{m}(s) \ge 0$ for all $(k, h, m, s) \in [K] \times [H] \times [M] \times \mathcal{S}$.

\paragraph{Step 1: Optimism.}
We want to show that $\Delta_{k,h}^{m}(s) = V_{k,h}^m(s) - V_h^{m, \star}(s) \ge 0$ for all $k, h, m, s$. We can prove this by induction on $h$.
For $h=H+1$, $V_{k,H+1}^m(s) = V_{H+1}^{m,\star}(s) = 0$ for all $m, k, s$. We can then use backward induction to show that the optimism holds for all $h \in [H]$.
For any $(s, h)$ at episode $k$ of agent $m$, denoting $a^\star=\pi_{h}^{m,\star}(s)$ for notational simplicity, since $Q_{k,h}^m=\min\{H-h+1,\ r_h^m+\hat P V_{k,h+1}^m+B_{k,h}^m\}$, we first prove that if the minimum in $Q_{k,h}^m(s,a^\star)$ is attained by $H-h+1$, then $Q_{k,h}^m(s,a^\star)=H-h+1\ge Q_h^{m,\star}(s,a^\star)$; otherwise, we can prove the optimism by the estimation error decomposition as follows:
\begin{align} \label{eq:fed_ucbvi_optimism_decomposition}
\Delta_{k,h}^{m}(s) = V_{k,h}^m(s) - V_h^{m, \star}(s)
&\stackrel{(i)}{\ge} Q_{k,h}^m(s, a^\star) - Q_h^{m, \star}(s, a^\star) \cr
&= \Big( r_{h}^m(s,a^\star) + \hat{P}_{k,h,s,a^\star} V_{k,h+1}^m + B_{k,h}^m(s,a^\star) \Big) \cr
&\quad - \Big( r_{h}^m(s,a^\star) + P_{h,s,a^\star} V_{h+1}^{m, \star} \Big) \cr
&= \big(\hat{P}_{k,h,s,a^\star} - P_{h,s,a^\star}\big) V_{k,h+1}^m + P_{h,s,a^\star} \big(V_{k,h+1}^m - V_{h+1}^{m, \star}\big) \cr
&\quad + B_{k,h}^m (s,a^\star) \cr
&\stackrel{(ii)}{\ge}- B_{k,h}^m (s,a^\star) - \frac{2}{H} P_{h,s,a^\star} \Delta_{k,h+1}^m \cr
&\quad + P_{h,s,a^\star} \Delta_{k,h+1}^m + B_{k,h}^m (s,a^\star) \cr
&= \left( 1 - \frac{2}{H} \right) P_{h,s,a^\star} \Delta_{k,h+1}^m \ge 0.
\end{align}
where (i) holds as $V_{k,h}^m(s)\ge Q_{k,h}^m(s,a^\star)$ and $V_h^{m,\star}(s)=Q_h^{m,\star}(s,a^\star)$ and (ii) follows from Lemma \ref{lemma:transition_error_bound} and the induction hypothesis $\Delta_{k,h+1}^m(s) \ge 0$ for all $s$; for $H=1$, the last line is $0$ since $\Delta^m_{k,2}=0$.

\begin{lemma}(Transition estimation error bound) \label{lemma:transition_error_bound} 
With probability at least $1-2\delta/7$, simultaneously for all episodes $k$, steps $h$, state-action pairs $(s,a)$, and agents $m$ such that $\Delta_{k,h+1}^m \ge 0$, we have
\begin{align}
-B_{k,h}^m(s,a) - \frac{2}{H} P_{h,s,a} \Delta_{k,h+1}^m \le \big(\hat{P}_{k,h,s,a} - P_{h,s,a}\big) V_{k,h+1}^m \le B_{k,h}^m(s,a) + \frac{2}{H} P_{h,s,a} \Delta_{k,h+1}^m
\end{align}
\end{lemma}

Once the optimism is established, for each agent $m$ at episode $k$, we can decompose the regret as follows:%
\begin{align} \label{eq:regret_optimism_final}
\mexp_{s_1\sim\rho} \left[V_1^{m, \star}(s_1) - V_1^{m, \pi_k^m}(s_1) \right]
&\le \mexp_{s_1\sim\rho} \left[V_{k, 1}^{m}(s_1) - V_1^{m, \pi_k^m}(s_1) \right] %
\end{align}

\paragraph{Step 2: Establishing the Recursion.}

To further bound $\Delta_{k,h}^{m,\pi_{k}^m} = V_{k,h}^m - V_{h}^{m, \pi_k^m}$ in \eqref{eq:regret_optimism_final}, for any step $h$ and state-action pair $(s_h, a_h)$ induced by $\rho$ and $a_h = \pi_{k,h}^m(s_h) = \argmax_{a \in \mathcal{A}} Q_{k,h}^{m}(s_h, a)$, we decompose the estimation error as follows:
\begin{align} \label{eq:regret_step_decomposition}
&\Delta_{k,h}^{m,\pi_{k}^m}(s_h) = V_{k,h}^m(s_h) - V_{h}^{m, \pi_k^m} (s_h) = Q_{k,h}^{m}(s_h, a_h) - Q_h^{m, \pi_k^m}(s_h, a_h) \cr
&\le \min \Big \{ H, r_{h}^m(s_h,a_h) + \hat{P}_{k,h,s_h,a_h} V_{k,h+1}^m + B_{k,h}^m(s_h,a_h) \ - \Big( r_{h}^m(s_h,a_h) + P_{h,s_h,a_h} V_{h+1}^{m, \pi_k^m} \Big) \Big \}\cr
&\le \min \Big \{ H, B_{k,h}^m(s_h,a_h) + \big(\hat{P}_{k,h,s_h,a_h} - P_{h,s_h,a_h}\big) V_{k,h+1}^m + P_{h,s_h,a_h} \Delta_{k,h+1}^{m,\pi_{k}^m} \Big \} \cr
&\stackrel{(i)}{\le} \min \Big \{ H, 2 B_{k,h}^m(s_h,a_h) + \frac{2}{H} P_{h,s_h,a_h} \Delta_{k,h+1}^{m} + P_{h,s_h,a_h} \Delta_{k,h+1}^{m,\pi_{k}^m} \Big \} \cr
&\stackrel{(ii)}{\le} \min  \{ H, 2 B_{k,h}^m(s_h,a_h) \}+ \left( 1 + \frac{2}{H} \right) P_{h,s_h,a_h} \Delta_{k,h+1}^{m,\pi_{k}^m}, %
\end{align}
where (i) holds due to Lemma \ref{lemma:transition_error_bound}, which yields:
\begin{align}
\big(\hat{P}_{k,h,s_h,a_h} - P_{h,s_h,a_h}\big) V_{k,h+1}^m \le B_{k,h}^m(s_h,a_h) + \frac{2}{H} P_{h,s_h,a_h} \Delta_{k,h+1}^{m} ,
\end{align}
(ii) holds because $V_{h+1}^{m, \star} \ge V_{h+1}^{m, \pi_k^m}$ and $\Delta_{k,h+1}^m \le \Delta_{k,h+1}^{m,\pi_{k}^m}$ according to the optimism shown in Step 1.

\paragraph{Step 3: Variance Transfer.}
Recall the definition of the exploration bonus $B_{k,h}^m(s,a)$, which utilizes the empirical variance of $V_{k,h+1}^m$:
\begin{align} \label{eq:bonus_empirical_def}
B_{k,h}^m(s,a) = \sqrt{ \frac{4 \Var_{\hat{P}_{k,h,s,a}}(V_{k,h+1}^m) \iota }{\max\{1,N_{k,h}(s,a)\}} } + \frac{25 S H^2 \iota}{ \max\{1, N_{k,h}(s,a)\}},
\end{align}
where $\iota$ is defined in the Notation. To bound the bonus term, we need to transfer the empirical variance of $V_k$ to that of $V^\star$, and then transfer the empirical variance $\Var_{\hat{P}}$ to the true variance $\Var_{P}$.

\begin{itemize}
    \item $\Var_{\hat P}(V_{k,h+1}^m)\to\Var_{\hat P}(V_{h+1}^{m,\star})$: First, we transfer the empirical variance of the estimated value function to that of the optimal value function. Writing $V_{k,h+1}^m = V_{h+1}^{m,\star} + \Delta_{k,h+1}^m$ with $\Delta_{k,h+1}^m = V_{k,h+1}^m - V_{h+1}^{m,\star}$, and denoting $\hat{P} = \hat{P}_{k,h,s,a}$ for brevity, we have
\begin{align}
\Var_{\hat{P}}(V_{k,h+1}^m)
&= \Var_{\hat{P}}\big(V_{h+1}^{m,\star} + \Delta_{k,h+1}^m\big) \cr
&= \mexp_{s' \sim \hat{P}}\Big[\Big(\big(V_{h+1}^{m,\star}(s') - \hat{P}V_{h+1}^{m,\star}\big) + \big(\Delta_{k,h+1}^m(s') - \hat{P}\Delta_{k,h+1}^m\big)\Big)^2\Big] \cr
&\overset{(i)}{\le} 2\,\mexp_{s' \sim \hat{P}}\Big[\big(V_{h+1}^{m,\star}(s') - \hat{P}V_{h+1}^{m,\star}\big)^2\Big]
+ 2\,\mexp_{s' \sim \hat{P}}\Big[\big(\Delta_{k,h+1}^m(s') - \hat{P}\Delta_{k,h+1}^m\big)^2\Big] \cr
&= 2\Var_{\hat{P}}(V_{h+1}^{m,\star}) + 2\Var_{\hat{P}}(\Delta_{k,h+1}^m) \cr
&\overset{(ii)}{=} 2\Var_{\hat{P}}(V_{h+1}^{m,\star}) + 2\Big(\hat{P}\big(\Delta_{k,h+1}^m\big)^2 - \big(\hat{P}\Delta_{k,h+1}^m\big)^2\Big) \cr
&\overset{(iii)}{\le} 2\Var_{\hat{P}}(V_{h+1}^{m,\star}) + 2\,\hat{P}\big(\Delta_{k,h+1}^m\big)^2 \cr
&\overset{(iv)}{\le} 2\Var_{\hat{P}}(V_{h+1}^{m,\star}) + 2H\,\hat{P}\Delta_{k,h+1}^m ,
\end{align}
where (i) follows from the elementary inequality $(x+y)^2 \le 2x^2 + 2y^2$ applied pointwise, (ii) is the definition of variance, (iii) drops the non-negative term $(\hat{P}\Delta_{k,h+1}^m)^2 \ge 0$, and (iv) holds since the optimism property (Step 1) and $V_{k,h+1}^m\le H$ and $V_{h+1}^{m,\star}\ge0$ guarantee $0 \le \Delta_{k,h+1}^m(s') \le H$ for all $s'$.

    \item $\Var_{\hat P}(V_{h+1}^{m,\star})\to\Var_P(V_{h+1}^{m,\star})$: Next, we convert the empirical variance of the optimal value function to the true variance by applying Lemma~\ref{lemma:ucbvi_variance_gap} as follows:
    \begin{align} \label{eq:var_gap_applied}
\Var_{\hat{P}}(V_{h+1}^{m,\star}) \le 2\Var_{P}(V_{h+1}^{m,\star}) + \frac{10 H^2 \iota}{\max\{N_{k,h}(s,a),1\}}.
\end{align}

\begin{lemma}(Empirical Variance Gap)\label{lemma:ucbvi_variance_gap}
With probability at least $1-2\delta/7$, for all $(k,h,s,a,m)$, the deviation between the empirical and true variance of the fixed function $V_{h+1}^{m,\star}\in[0,H]$ is bounded by:
\begin{align} \label{eq:variance_concentration}
\big| \Var_{\hat{P}_{k,h, s,a}}(V_{h+1}^{m, \star}) - \Var_{P_{h,s,a}}(V_{h+1}^{m, \star}) \big| \le \Var_{P_{h,s,a}}(V_{h+1}^{m, \star}) + \frac{10 H^2 \iota}{\max\{ N_{k,h}(s,a),1\}}.
\end{align}
\end{lemma}

\item \textbf{Bound on $\hat{P}\Delta_{k,h+1}^m$}:
Invoking the bound on $|(\hat{P} - P)\Delta_{k,h+1}^m|$ established in the proof of Lemma~\ref{lemma:transition_error_bound}, we have
\begin{align} \label{eq:phat_delta_to_p_delta}
\hat{P}\Delta_{k,h+1}^m 
\le P\Delta_{k,h+1}^m + \big|(\hat{P} - P)\Delta_{k,h+1}^m\big|
&\le \Big(1 + \frac{1}{H}\Big) P\Delta_{k,h+1}^m + \frac{9 S H^2 \iota}{\max\{N_{k,h}(s,a),1\}} \cr
&\le 2 P\Delta_{k,h+1}^m + \frac{9 S H^2 \iota}{\max\{N_{k,h}(s,a),1\}}.
\end{align}

\end{itemize}

Combining the variance transfer above, \eqref{eq:var_gap_applied} and \eqref{eq:phat_delta_to_p_delta},
\begin{align}
\Var_{\hat{P}}(V_{k,h+1}^m) 
\le 4\Var_{P}(V_{h+1}^{m,\star}) + 4H \, P_{h,s,a}\Delta_{k,h+1}^m + \frac{38\, S H^3 \iota}{ \max\{N_{k,h}(s,a),1\}}.
\end{align}
Substituting this into the bonus definition \eqref{eq:bonus_empirical_def},
\begin{align}
B_{k,h}^m(s,a) 
&\le \sqrt{ \frac{16 \Var_{P_{h,s,a}}(V_{h+1}^{m,\star}) \iota }{\max\{N_{k,h}(s,a),1\} } }
+ \sqrt{ \frac{16 H \iota \cdot P_{h,s,a}\Delta_{k,h+1}^m}{ \max\{N_{k,h}(s,a),1\} } } \nonumber\\
&\quad+ \frac{13 \sqrt{S} H^{3/2} \iota}{\max\{N_{k,h}(s,a),1\} }
+ \frac{25 S H^2 \iota}{\max\{N_{k,h}(s,a),1\} }.
\end{align}
For the second term, applying Young's inequality $\sqrt{xy} \le \alpha x + \frac{y}{4\alpha}$ with $\alpha = \frac{1}{4e^2 H}$, $x = P_{h,s,a}\Delta_{k,h+1}^m$, and $y = \frac{16 H \iota}{\max\{N_{k,h}(s,a),1\}}$:
\begin{align}
\sqrt{ \frac{16 H \iota \cdot P_{h,s,a}\Delta_{k,h+1}^m}{\max\{N_{k,h}(s,a),1\} } }
\le \frac{1}{4e^2 H} P_{h,s,a}\Delta_{k,h+1}^m + \frac{16 e^2 H^2 \iota}{\max\{N_{k,h}(s,a),1\} }.
\end{align}
Since $S,H\ge1$, the three lower-order terms satisfy $13\sqrt{S}H^{3/2}+25SH^2+16e^2H^2\le157SH^2$, and we arrive at
\begin{align} \label{eq:bonus_true_transition}
B_{k,h}^m(s,a) &\le \underbrace{\sqrt{ \frac{16 \Var_{P_{h,s,a}}(V_{h+1}^{m, \star}) \iota }{\max\{N_{k,h}(s,a),1\} } } + \frac{157 S H^2 \iota}{\max\{N_{k,h}(s,a),1\} }}_{=: \tilde{B}_{k,h}^m(s,a)} + \frac{1}{4e^2 H} P_{h,s,a} \Delta_{k,h+1}^m \cr
&\le \tilde{B}_{k,h}^m(s,a) + \frac{1}{4e^2 H} P_{h,s,a} \Delta_{k,h+1}^{m,\pi_k^m},
\end{align}
defining the residual-free bonus $\tilde{B}_{k,h}^m(s,a)$ that depends only on the true variance of the optimal value function $V_{h+1}^{m, \star}$ and the last term holds since $\Delta_{k,h+1}^m \le \Delta_{k,h+1}^{m,\pi_k^m}$ by optimism (Step 1).

\paragraph{Step 4: Bounding the Total Regret via Martingale Concentration.}
Integrating the variance transfer \eqref{eq:bonus_true_transition} from Step 3 into the estimation error bound \eqref{eq:regret_step_decomposition} of Step 2, substituting the bonus term $B_{k,h}^m$ with $\tilde{B}_{k,h}^m$, we have
\begin{align}
\Delta_{k,h}^{m,\pi_{k}^m}(s_h) 
&\le \min  \{ H, 2 \tilde B_{k,h}^m(s_h,a_h) \}+ \left( 1 + \frac{2}{H} \right) P_{h,s_h,a_h} \Delta_{k,h+1}^{m,\pi_{k}^m} + \frac{1}{2e^2 H} P_{h,s_h,a_h} \Delta_{k,h+1}^{m,\pi_k^m}\cr
&\le \min  \{ H, 2 \tilde B_{k,h}^m(s_h,a_h) \}+ \left( 1 + \frac{3}{H} \right) P_{h,s_h,a_h} \Delta_{k,h+1}^{m,\pi_{k}^m} .
\end{align}

For the realized trajectory of agent $m$ in episode $k$, let $h \in [H]$ and $s^m_{k,h}, a^m_{k,h}$ be the state and action at time $h$ and $P=P_{h,s^m_{k,h},a^m_{k,h}}$,
\begin{align} \label{eq:realized_recursion}
\Delta^{m,\pi_k^m}_{k,h}(s^m_{k,h})
&\le\min\{H,2\tilde B^m_{k,h}(s^m_{k,h},a^m_{k,h})\} \cr
&\quad +\Big(1+\frac3H\Big)\Big(\Delta^{m,\pi_k^m}_{k,h+1}(s^m_{k,h+1})+ \underbrace{(P_{h,s^m_{k,h},a^m_{k,h}} \Delta^{m,\pi_k^m}_{k,h+1} - \Delta^{m,\pi_k^m}_{k,h+1}(s^m_{k,h+1}))}_{\defn \xi_{k,h}^m } \Big),
\end{align}
where $\xi_{k,h}^m$ is a martingale difference, which has conditional mean zero given the past and $(s_{k,h}^m,a_{k,h}^m)$, and $\xi_{k,h}^m$ lies in an interval of length $2H$ given the past, since $V^m_{k,h+1},V^{m,\pi_k^m}_{h+1}\in[0,H]$.

Unrolling the recursion along the realized trajectory from $h=1$ to $H$, with $\Delta^{m,\pi_k^m}_{k,H+1}=0$, and then using $(1+3/H)^{h-1}\le e^3$ and $\min\{H,2x\}\le2\min\{H,x\}$, 
\begin{align}
\Delta^{m,\pi_k^m}_{k,1}(s^m_{k,1})&\le\sum_{h=1}^H\Big(1+\frac3H\Big)^{h-1}\min\{H,2\tilde B^m_{k,h}(s^m_{k,h},a^m_{k,h})\}+\sum_{h=1}^H\Big(1+\frac3H\Big)^h\xi^m_{k,h} \nonumber \\
&\le 2e^3\sum_{h=1}^H\min\{H,\tilde B^m_{k,h}(s^m_{k,h},a^m_{k,h})\}+\sum_{h=1}^H\Big(1+\frac3H\Big)^h\xi^m_{k,h}.
\end{align}

We now bound the total regret by summing over all agents $m \in [M]$ and all episodes $k \in [K]$ using the bounds from the previous steps, and applying the martingale difference sequence (MDS) concentration argument to the sum of martingale differences $\xi_{k,h}^m$:
\begin{align}
\mathsf{Regret}(K)
&\le \sum_{m=1}^M\sum_{k=1}^K\mexp_{s_1\sim\rho}\big[\Delta^{m,\pi_k^m}_{k,1}(s_1)\big] \nonumber \\
&= \sum_{m=1}^M\sum_{k=1}^K\Big(\Delta^{m,\pi_k^m}_{k,1}(s^m_{k,1})+\underbrace{\mexp_{s_1\sim\rho}[\Delta^{m,\pi_k^m}_{k,1}(s_1)]-\Delta^{m,\pi_k^m}_{k,1}(s^m_{k,1})}_{=: \zeta^m_k} \Big) \nonumber\\
&\le 2e^3\sum_{m,k,h}\min\{H,\tilde B^m_{k,h}(s^m_{k,h},a^m_{k,h})\}
+\sum_{m=1}^M\sum_{k=1}^K\Big(\zeta^m_k+\sum_{h=1}^H\Big(1+\frac3H\Big)^h\xi^m_{k,h}\Big)\nonumber \\
&= 2e^3\sum_{m,k,h}\min\{H,\tilde B^m_{k,h}(s^m_{k,h},a^m_{k,h})\}+\sum_{\tau=1}^{MK(H+1)}X_\tau, \nonumber %
\end{align}
where $X_\tau=\zeta^m_k$ for $h=0$, $X_\tau=(1+3/H)^h\xi^m_{k,h}$ for $h\ge1$ for each $\tau$ corresponding to exactly one $(k,m,h)$ with $h\in\{0,\dots,H\}$. 

To apply the Azuma-Hoeffding inequality to $\{X_\tau\}$, we order the steps of all agents and episodes. We define a global rollout index $\tau \in [1, MK(H+1)]$ that strictly flattens the multi-agent episodes into a step-by-step sequential timeline:
\begin{align}
\tau = (k-1)M(H+1)+(m-1)(H+1)+h+1 \quad \text{for} \quad k \in [K], \ m \in [M], \ h\in\{0,\dots,H\}.
\end{align}
Let $\mathcal{F}_\tau$ be the filtration (a $\sigma$-algebra) capturing all information up to and including the state-action pair of the $\tau$-th step; the slot $h=0$ carries no state-action pair. Formally, $\mathcal{F}_\tau$ is defined as:
\begin{align}
\mathcal{F}_\tau = \sigma \left( \bigcup_{i=1}^{\tau} \big\{ s^{(i)}, a^{(i)} \big\} \right), \quad \mathcal{F}_1 = \{\emptyset, \Omega\},
\end{align}
where the superscript $(i)$ refers to the state-action pair at the specific global step index $i$. Since the global estimator $\hat{P}$ and the counter $N$ are updated strictly using past data contained in the episode-start filtration $\mathcal{F}_{(k-1)M(H+1)+1} \subset \mathcal{F}_\tau$, the policy $\pi_k^m$ and the bonus function $\tilde{B}_{k,h}^m$ evaluated during step $\tau$ are $\mathcal{F}_\tau$-measurable. %

As $X_\tau$ is $\mathcal F_{\tau+1}$-measurable, with $\mexp[X_\tau\mid\mathcal F_\tau]=0$ and $X_\tau$ lying in an $\mathcal F_\tau$-measurable interval of length $2e^3H$, applying the Azuma-Hoeffding inequality to the bounded MDS $\sum_{\tau=1}^{MK(H+1)}X_\tau$, we guarantee that with probability at least $1-\delta/7$:
\begin{align} \label{eq:mds_concentration}
\sum_{\tau=1}^{MK(H+1)} X_\tau \le \sqrt{\frac{1}{2} \sum_{\tau=1}^{MK(H+1)} \big( 2e^3H \big)^2 \iota} \le 2e^3 \sqrt{ H^3 M K \iota}.
\end{align}

Substituting the concentration bound \eqref{eq:mds_concentration} into the total expected regret, we have
\begin{align}
\mathsf{Regret}(K)
&\le \sum_{m=1}^M\sum_{k=1}^K\mexp_{s_1\sim\rho}\big[\Delta^{m,\pi_k^m}_{k,1}(s_1)\big] \nonumber\\
&\le 2e^3\sum_{m,k,h}\min\{H,\tilde B^m_{k,h}(s^m_{k,h},a^m_{k,h})\}+2e^3\sqrt{H^3MK\iota},
\end{align}

\paragraph{Step 5: Finalizing the Regret Bound.}
Now, we substitute the upper bound of the empirical exploration bonus into the total expected regret. 
\begin{align} \label{eq:total_regret_empirical}
\mathsf{Regret}(K)
&\le 2e^3\sum_{m,k,h}\min\{H,\tilde B^m_{k,h}(s^m_{k,h},a^m_{k,h})\} + 2e^3\sqrt{H^3 M K \iota}  \cr
&\le 4e^3MSAH^2+2e^3\sum_{N_{k,h}\ge M}\tilde B^m_{k,h}(s^m_{k,h},a^m_{k,h}) + 2e^3\sqrt{H^3 M K \iota}  \cr
&\le 4e^3MSAH^2+8e^3\sum_{N_{k,h}\ge M} \sqrt{ \frac{ \Var_{P_{h,s_{k,h}^m, a_{k,h}^m}}(V_{h+1}^{m, \star}) \iota }{\max\{N_{k,h}(s^m_{k,h},a^m_{k,h}),1\} } } \cr
&\quad+ 314e^3 \sum_{N_{k,h}\ge M} \frac{ S H^2 \iota}{\max\{N_{k,h}(s^m_{k,h},a^m_{k,h}),1\} } + 2e^3\sqrt{H^3 M K \iota} \cr
&\le 8e^3 \sqrt{\iota } \underbrace{\sqrt{\sum_{m,k,h} \Var_{P_{h,s_{k,h}^m, a_{k,h}^m}}(V_{h+1}^{m, \star})(s_{k,h}^m, a_{k,h}^m) }}_{\text{(I)}} \cdot \sqrt{ \sum_{N_{k,h}\ge M}\frac{1}{\max\{N_{k,h}(s^m_{k,h},a^m_{k,h}),1\} } }  \cr
&\quad+4e^3MSAH^2+ 314e^3 S H^2  \iota \underbrace{\sum_{N_{k,h}\ge M} \frac{1}{\max\{N_{k,h}(s^m_{k,h},a^m_{k,h}),1\} }}_{\text{(II)}} + 2e^3\sqrt{H^3 M K \iota} ,
\end{align}
where $\sum_{N_{k,h}\ge M}$ denotes the sum over $(m,k,h)$ with $N_{k,h}(s^m_{k,h},a^m_{k,h})\ge M$. The second inequality uses $\min\{H,\tilde B\}\le H$ for the visits with $N_{k,h}(s^m_{k,h},a^m_{k,h})<M$; for each $(h,s,a)$, there are fewer than $2M$ such visits, since the counter increases by at most $M$ per episode. The last inequality uses the Cauchy-Schwarz inequality.

\begin{itemize}
    \item \textbf{Bound of term (I):} The term (I) is the sum of the variances of the value functions, with $P=P_{h,s^m_{k,h},a^m_{k,h}}$. Since $0\le V^{m,\pi_k^m}_{h+1}\le V^{m,\star}_{h+1}\le H$, as in \citet[Lemma~9]{azar2017minimax},
    \begin{align} \label{eq:total_var_split}
    \Var_P(V^{m,\star}_{h+1})-\Var_P(V^{m,\pi_k^m}_{h+1})\le P\big((V^{m,\star}_{h+1})^2-(V^{m,\pi_k^m}_{h+1})^2\big)\le2H\,P\Delta^{m,\pi_k^m}_{k,h+1}.
    \end{align}
    Hence $\text{(I)}^2\le\sum_{m,k,h}\Var_P(V^{m,\pi_k^m}_{h+1})+2H\sum_{m,k,h}P\Delta^{m,\pi_k^m}_{k,h+1}$, and we bound the two sums in turn.
\begin{itemize}
\item \textbf{Bounding $\sum_{m,k,h}\Var_P(V^{m,\pi_k^m}_{h+1})$}: For each $(m,k)$, conditionally on the history before the episode $k$ of agent $m$, the law of total variance \citep[Lemma~8]{azar2017minimax} gives $\mexp[\sum_{h=1}^H\Var_P(V^{m,\pi_k^m}_{h+1})]\le H^2$; the trajectories of the other agents in episode $k$ do not affect that of agent $m$ given $\pi_k^m$. Since $0\le\sum_{h=1}^H\Var_P(V^{m,\pi_k^m}_{h+1})\le H^3$, the Azuma-Hoeffding inequality over the $MK$ pairs $(m,k)$ gives, with probability at least $1-\delta/7$,
    \begin{align} \label{eq:total_variance_bound}
    \sum_{m,k,h}\Var_P(V_{h+1}^{m,\pi_k^m})\le MKH^2+H^3\sqrt{2MK\iota}.
    \end{align}

 \item   \textbf{Bounding $\sum_{m,k,h}P\Delta^{m,\pi_k^m}_{k,h+1}$}: As in \citet[Lemma~7]{azar2017minimax}, write $P\Delta^{m,\pi_k^m}_{k,h+1}=\Delta^{m,\pi_k^m}_{k,h+1}(s^m_{k,h+1})+\xi^m_{k,h}$ and unroll \eqref{eq:realized_recursion} from step $h+1$:
    \begin{align}
    P\Delta^{m,\pi_k^m}_{k,h+1}\le2e^3\sum_{i=1}^H\min\{H,\tilde B^m_{k,i}(s^m_{k,i},a^m_{k,i})\}+\sum_{i=h}^H\Big(1+\frac3H\Big)^{i-h}\xi^m_{k,i}.
    \end{align}
    Sum over $(m,k,h)$. Each $\xi^m_{k,i}$ then appears with total weight $\sum_{h\le i}(1+3/H)^{i-h}\le e^3H$, so the second term is a sum of $MKH$ martingale differences each lying in an interval of length $2e^3H^2$ given the past. By the Azuma-Hoeffding inequality, it is at most $e^3H^2\sqrt{2MKH\iota}$ with probability at least $1-\delta/7$. For the first term, the same steps as in \eqref{eq:total_regret_empirical}, including the split at $N_{k,h}<M$, and the bound \eqref{eq:harmonic_sum} on (II) give
    \begin{align}
    \sum_{m,k,h}\min\{H,\tilde B^m_{k,h}(s^m_{k,h},a^m_{k,h})\}\le4\sqrt{2SAH\iota\log K}\cdot\text{(I)}+\mathcal{O}\big(MSAH^2+S^2AH^3\iota\log K\big),
    \end{align}
    where $\mathcal{O}(\cdot)$ hides absolute constants only. Therefore,
    \begin{align} \label{eq:residual_sum_bound}
    \sum_{m,k,h}P\Delta^{m,\pi_k^m}_{k,h+1}
    &\le8e^3H\sqrt{2SAH\iota\log K}\cdot\text{(I)} \\
    &\quad+e^3\sqrt{2H^5MK\iota}+\mathcal{O}\big(MSAH^3+S^2AH^4\iota\log K\big). \nonumber
    \end{align}

  \item  \textbf{Combining.} Substitute \eqref{eq:total_variance_bound} and \eqref{eq:residual_sum_bound}. The square-root terms in \eqref{eq:total_variance_bound} and \eqref{eq:residual_sum_bound} contribute $\mathcal{O}(MKH^2+H^5\iota)$ by the AM-GM inequality. Hence
    \begin{align} \label{eq:total_variance_final}
    \text{(I)}^2\le16e^3H^2\sqrt{2SAH\iota\log K}\cdot\text{(I)}+\mathcal{O}\big(MKH^2+MSAH^4+S^2AH^5\iota\log (4K)\big),
    \end{align}
    and solving this quadratic inequality in (I),%
    \begin{align}
\text{(I)}=\mathcal{O}\Big(\sqrt{MKH^2}+\sqrt{MSAH^4}+\sqrt{S^2AH^5\iota^2}\Big).
    \end{align}

\end{itemize}

    \item \textbf{Bound of term (II):} The counter $N_{k,h}(s,a)$ is shared by all agents and fixed during episode $k$, and $N_{k+1,h}(s,a)=N_{k,h}(s,a)+\sum_{m=1}^M\ind\{(s^m_{k,h},a^m_{k,h})=(s,a)\}$. Hence,
    \begin{align} \label{eq:harmonic_sum}
    \text{(II)}
    &=\sum_{k,h}\ \sum_{(s,a):\,N_{k,h}(s,a)\ge M}\frac{\sum_{m=1}^M\ind\{(s^m_{k,h},a^m_{k,h})=(s,a)\}}{N_{k,h}(s,a)}\cr
    &=\sum_{h,s,a}\ \sum_{k:\,N_{k,h}(s,a)\ge M}\frac{N_{k+1,h}(s,a)-N_{k,h}(s,a)}{N_{k,h}(s,a)}\cr
    &\overset{(i)}{\le}2\sum_{h,s,a}\ \sum_{j=M+1}^{MK}\frac1j
    \le2SAH\log K.
    \end{align}
    In (i), $N_{k,h}(s,a)\ge M$ implies $N_{k+1,h}(s,a)\le2N_{k,h}(s,a)$, so each fraction is at most $2\sum_{j=N_{k,h}(s,a)+1}^{N_{k+1,h}(s,a)}1/j$, and these ranges of $j$ are disjoint over $k$ and lie in $\{M+1,\dots,MK\}$.

\end{itemize}

Substituting these bounds on (I) and (II) into \eqref{eq:total_regret_empirical}, with probability at least $1-\delta$,
\begin{align}
\mathsf{Regret}(K)=\mathcal{O}\Big(\sqrt{H^3SAMK\iota^2}+S^2AH^3\iota^2+MSAH^2\Big).
\end{align}

\providecommand{\wfix}[2]{{\color{orange}\textbf{[Weighted fix-#1]}~#2}}

\section{Proof of Theorem~\ref{thm:main_regret}}
\label{sec:proof-fedefvi-regret}

The proof proceeds in four major steps, demonstrating how the federated global coverage eliminates the need for action-wise exploration bonuses. Throughout this proof, $(h,s,a)$ ranges over $s\in\mathcal S_h$; this is without loss of generality since, for $N_{k,h}(s,a)\ge1$, $\hat P_{k,h}(\cdot|s,a)$ is supported on observed next states, which lie in $\mathcal S_{h+1}$.

\paragraph{Step 1: Global Concentration via Reward Heterogeneity. }
Under Assumption~\ref{assumption:expfree_global_coverage}, the heterogeneity of the agents' rewards guarantees that their greedy policies under the estimate $\hat P_k$ used by the algorithm jointly cover the state-action space. Although individual agents act purely greedily, the federated global count $N_{k,h}(s, a)$ grows linearly with the episode $k$.

\begin{lemma}[Global concentration via heterogeneity]
\label{lemma:global_concentration}
Suppose Assumption \ref{assumption:expfree_global_coverage} holds. Let $N_{k,h}(s, a)$ be the global visitation count for any state-action pair $(s,a) \in \mathcal{S} \times \mathcal{A}$ at step $h \in [H]$ until episode $k$. {With probability at least $1-\frac{\delta}{2}$, for all $k\ge1$, $h\in[H]$, and $(s,a)\in\mathcal S_h\times\mathcal A$:
\begin{align}
N_{k+1,h}(s,a)\ge\frac12(k-1)M\lambda_{\mathsf{min}}-\ln\frac{2SAH}{\delta}.
\end{align}
Consequently, with the burn-in threshold $K_{th}=\big\lceil 4\log(4M^2SAKH/\delta)/(M\lambda_{\mathsf{min}})\big\rceil+2$, for all $k\ge K_{th}$ the global visitation count is lower bounded by a linear rate: $N_{k+1,h}(s,a)\ge\frac14kM\lambda_{\mathsf{min}}$.}
\end{lemma}

This implies that the environment is thoroughly explored by the collective behaviors of the heterogeneous agents, completely bypassing the need for forced exploration.

\paragraph{Step 2: Shrinking Bellman Error. }
With the linearly growing global visitation counts for $k > K_{th}$, the empirical transition dynamics $\hat{P}_{k,h}$ rapidly converge to the true dynamics $P_h$ as follows:

\begin{lemma}[Shrinking Bellman Error]
\label{lemma:shrinking_error}
Suppose the result of Lemma \ref{lemma:global_concentration} holds. Let $Q_{k,h}^m(s, a)$ be the action-value function computed by agent $m$ at episode $k$ using the empirical transition $\hat{P}_{k,h}$, and let $Q_h^{m, \star}(s, a)$ be the true optimal action-value function. With probability at least $1 - \delta$, for all $k > K_{th}$, $m \in [M]$, $h \in [H]$, and $(s,a) \in \mathcal{S}_h \times \mathcal{A}$:
\begin{align}
    \left| Q_{k,h}^m(s, a) - Q_h^{m, \star}(s, a) \right| \le  \mathcal{O}  \left( \sqrt{\frac{ H^4 \log(4 K M H S A / \delta) }{ M \lambda_{\mathsf{min}} k}} + \frac{H^2 \log(4 K M H S A / \delta)}{M \lambda_{\mathsf{min}} k} \right )
\end{align}
\end{lemma}

\paragraph{Step 3: Burn-in Costs. }
Because PF-EFVI does not inject any UCB bonuses, an agent $m$ will strictly choose the optimal action $\pi_h^{m, \star}(s)$ if its Q-value estimation error is strictly less than half of the minimum suboptimality gap, i.e., $\left| Q_{k,h}^m - Q_h^{m, \star} \right| < \Delta_{\min}^m / 2$. We derive the critical burn-in episode $K_0^m$ for agent $m$ as follows:

\begin{lemma}[The burn-in cost of pure exploitation]
\label{lemma:tipping_point}
There exists a critical burn-in episode $K_0^m$ such that for all episodes $k > \max(K_{th}, K_0^m)$, the purely greedy policy $\pi_{k,h}^m(s) = \arg\max_a Q_{k,h}^m(s, a)$ selects an optimal action at every state, yielding zero instantaneous regret. This tipping point is bounded by:
\begin{align}
    K_0^m = \mathcal{O} \left( \frac{H^4 \log(K M H S A / \delta)}{M \lambda_{\mathsf{min}} (\Delta_{\min}^m)^2} \right)
\end{align}
\end{lemma}

\paragraph{Step 4: Summing the Regret. }
To compute the total regret for agent $m$, we split the learning process into the initial uncontrolled phase ($k \le K_{th}$) and the shrinking phase ($K_{th} < k \le K_0^m$).
If a suboptimal action is chosen, its gap is bounded by $2 \epsilon_k$. 
Since the instantaneous regret is strictly zero for all $k > \max(K_{th}, K_0^m)$ (by Lemma \ref{lemma:tipping_point}), we sum the regret across these two phases.

During the uncontrolled phase ($P1$), the instantaneous regret is trivially bounded by the horizon $H$. Thus:
\begin{align}
    \mathsf{Regret}^m_{P1}(K) \le \sum_{k=1}^{K_{th}} H = H \cdot K_{th}\label{eq:p1} \le \frac{4H\log(4M^2SAKH/\delta)}{M\lambda_{\mathsf{min}}}+3H
\end{align}

During the shrinking phase ($P2$), we integrate the bounded error $\epsilon_k$:
\begin{align}
    \mathsf{Regret}^m_{P2}(K)
    &= \sum_{k=K_{th}+1}^{K_0^m} \mathbb{E}_{s_1 \sim \rho} \left [ V_1^{m, \star} (s_1) - V_1^{m, \pi_k^{m} } (s_1)  \right] \cr
    &\stackrel{(i)}{=} \sum_{k=K_{th}+1}^{K_0^m} \mathbb{E}_{\pi_k^m, s_1 \sim \rho} \left[ \sum_{h=1}^H \left( Q_h^{m, \star}(s_h, \pi_h^{m, \star}(s_h)) - Q_h^{m, \star}(s_h, \pi_{k,h}^m(s_h)) \right) \right] \cr
    &\le \sum_{k=1}^{K_0^m} \mathcal{O} \left( \sqrt{\frac{H^6 \log(K M H S A / \delta)}{k M \lambda_{\mathsf{min}}}} \right) \cr
    &\le \mathcal{O} \left( \sqrt{\frac{H^6 \log(K M H S A / \delta)}{M \lambda_{\mathsf{min}}}} \sum_{k=1}^{K_0^m} \frac{1}{\sqrt{k}} \right) \cr
    &= \mathcal{O} \left( \sqrt{\frac{H^6 \log(K M H S A / \delta)}{M \lambda_{\mathsf{min}}}} \cdot 2 \sqrt{K_0^m} \right) \label{eq:integral_cancellation}
\end{align}
where (i) follows from the Performance Difference Lemma. Since the estimated Q-value error at any step is bounded by $\epsilon_k$, the suboptimality gap of the chosen greedy action at each step is at most $2\epsilon_k$. Summing this over the $H$ steps of the trajectory yields an instantaneous regret bound of $2 H \epsilon_k$.

We substitute the {explicit} value of $\sqrt{K_0^m} = \mathcal{O} \left( \frac{H^2 \sqrt{\log(K M H S A / \delta)}}{\sqrt{M \lambda_{\mathsf{min}}} \Delta_{\min}^m} \right)$ into Equation (\ref{eq:integral_cancellation}):
\begin{align}
    \mathsf{Regret}^m_{P2}(K) &\le \mathcal{O} \left(  \sqrt{\frac{H^6 \log(K M H S A / \delta)}{M \lambda_{\mathsf{min}}}} \cdot \frac{H^2 \sqrt{\log(K M H S A / \delta)}}{\sqrt{M \lambda_{\mathsf{min}}} \Delta_{\min}^m} \right) \nonumber \\
    &= \mathcal{O} \left( \frac{H^5 \log(K M H S A / \delta)}{M \lambda_{\mathsf{min}} \Delta_{\min}^m} \right)
\end{align}

Summing these independent bounds over all $M$ agents yields the final total system regret:
\begin{align}
    \mathsf{Regret}(K) &= \sum_{m=1}^M \left( \mathsf{Regret}^m_{P1}(K) + \mathsf{Regret}^m_{P2}(K) \right) \nonumber \\
    &\le \sum_{m=1}^M\Big(\frac{4H\log(4M^2SAKH/\delta)}{M\lambda_{\mathsf{min}}}+3H\Big) + \sum_{m=1}^M \mathcal{O} \left( \frac{H^5 \log(K M H S A / \delta)}{M \lambda_{\mathsf{min}} \Delta_{\min}^m} \right) \nonumber \\
    &= \mathcal{O} \left( \frac{H \log(4 SAKMH / \delta)}{\lambda_{\mathsf{min}}} + MH \right) + \mathcal{O} \left( \frac{H^5 \log(K M H S A / \delta)}{M \lambda_{\mathsf{min}}} \sum_{m=1}^M \frac{1}{\Delta_{\min}^m} \right).
\end{align}

{\section{A visitation-weighted form of Theorem~\ref{thm:main_regret}}
\label{sec:proof-fedefvi-weighted}

In Theorem~\ref{thm:main_regret}, the coverage enters only through $\lambda_{\mathsf{min}}$, that is, through the worst-covered state-action pair. We show that the leading, gap-dependent term only needs coverage along the trajectories that matter for each agent.

\begin{proof}[Proof of Proposition~\ref{prop:wt_regret}]
We work on the events of Lemmas~\ref{lemma:global_concentration} and~\ref{lemma:shrinking_error} and of \eqref{eq:wt_l1} below, which hold simultaneously with probability at least $1-\delta$. We write $\lambda_{h'}$ for $\lambda_{h'}(s_{h'},a_{h'})$ inside trajectory expectations.

\paragraph{Step 1: where $\lambda_{\mathsf{min}}$ enters.} In Step~4 of the proof of Theorem~\ref{thm:main_regret}, the regret of each episode $k\in(K_{th},K_0^m]$ is bounded by $2H\epsilon_k$, where $\epsilon_k=\max_{h,s,a}|Q^m_{k,h}(s,a)-Q^{m,\star}_h(s,a)|$ is governed by the worst-covered pair through Lemma~\ref{lemma:shrinking_error}. We keep the rest of Step~4 and replace only this bound.

\paragraph{Step 2: regret decomposition.} Fix $m$ and $k>K_{th}$. By step (i) of \eqref{eq:integral_cancellation},
\begin{align}
\mexp_{s_1\sim\rho}\big[V^{m,\star}_1(s_1)-V^{m,\pi^m_k}_1(s_1)\big]=\sum_{h=1}^H\mexp_{(s,a)\sim d_h^{\pi^m_k}}\big[\Delta^m_h(s,a)\big].\label{eq:wt_pdl}
\end{align}
If $d_h^{\pi^m_k}(s,a)>0$, then $\pi^m_{k,h}(a|s)>0$, so $a\in\argmax_{a'}Q^m_{k,h}(s,a')$ and, with $a^\star=\pi^{m,\star}_h(s)$,
\begin{align}
\Delta^m_h(s,a)\le\big[Q^{m,\star}_h-Q^m_{k,h}\big](s,a^\star)+\big[Q^m_{k,h}-Q^{m,\star}_h\big](s,a).\label{eq:wt_split}
\end{align}
Thus only the underestimation at the optimal action and the overestimation at the played action matter. Lemma~\ref{lemma:shrinking_error} bounds both by $\epsilon_k$.

\paragraph{Step 3: one-sided errors.} Lemma~\ref{lemma:wt_pair_error} below bounds the overestimation along $\pi^m_k$ and the underestimation along $\pi^{m,\star}$, with the local error at each pair governed by its own coverage. Applying Lemma~\ref{lemma:wt_pair_error} to \eqref{eq:wt_split}, and using $\mexp^{\pi^{m,\star}}[\,\cdot\mid s_h=s,a_h=a^\star]=\mexp^{\pi^{m,\star}}[\,\cdot\mid s_h=s]$,
\begin{align}
\Delta^m_h(s,a)&\le CH\sqrt{\frac{\iota}{kM}}\Big(\mexp^{\pi^m_k}\Big[\sum_{h'=h}^H\lambda_{h'}^{-1/2}\,\Big|\,s_h=s,a_h=a\Big]\nonumber\\
&\quad+\mexp^{\pi^{m,\star}}\Big[\sum_{h'=h}^H\lambda_{h'}^{-1/2}\,\Big|\,s_h=s\Big]\Big) +\frac{CH^3\sqrt S\,\iota}{kM\lambda_{\mathsf{min}}}.\label{eq:wt_onesided}
\end{align}

\paragraph{Step 4: from pairs to $\Lambda^m$.} If the first term of \eqref{eq:wt_onesided} is at least the second, then $\Delta^m_h(s,a)$ is at most twice the first term, so $\Delta^m_h(s,a)\le(2\cdot\text{first term})^2/\Delta^m_h(s,a)$. We then use $(x+y)^2\le2x^2+2y^2$, the bound $(\mexp^{\pi}[\sum_{h'=h}^H\lambda_{h'}^{-1/2}])^2\le H\,\mexp^{\pi}[\sum_{h'=h}^H\lambda_{h'}^{-1}]$ (Jensen and Cauchy--Schwarz), and $\Delta^m_h(s,a)\ge\Delta^m_{\min}$. Otherwise, $\Delta^m_h(s,a)$ is at most twice the second term. In both cases, for every $(s,a)$ with $\pi^m_{k,h}(a|s)>0$ and $\Delta^m_h(s,a)>0$,
\begin{align}
\Delta^m_h(s,a)&\le\frac{CH^3\iota}{kM\Delta^m_{\min}}\Big(\mexp^{\pi^m_k}\Big[\sum_{h'=h}^H\frac1{\lambda_{h'}}\,\Big|\,s_h=s,a_h=a\Big]\nonumber\\
&\quad+\mexp^{\pi^{m,\star}}\Big[\sum_{h'=h}^H\frac1{\lambda_{h'}}\,\Big|\,s_h=s\Big]\Big) +\frac{CH^3\sqrt S\,\iota}{kM\lambda_{\mathsf{min}}}.\label{eq:wt_ident}
\end{align}
We substitute this into \eqref{eq:wt_pdl}. Pairs with $\Delta^m_h(s,a)=0$ contribute nothing, and the right-hand side is nonnegative, so the restriction $\Delta^m_h(s,a)>0$ can be dropped. The second expectation depends on $(s,a)$ only through $s$. For the first, if $(s_h,a_h)\sim d_h^{\pi^m_k}$ and the trajectory continues with $\pi^m_k$, then $(s_{h'},a_{h'})\sim d_{h'}^{\pi^m_k}$ by the Markov property, so
\[
\sum_{h=1}^H\sum_{h'=h}^H\mexp_{(s,a)\sim d_{h'}^{\pi^m_k}}\Big[\frac{1}{\lambda_{h'}(s,a)}\Big]=\sum_{h'=1}^Hh'\,\mexp_{(s,a)\sim d_{h'}^{\pi^m_k}}\Big[\frac{1}{\lambda_{h'}(s,a)}\Big]\le H\sum_{h=1}^H\mexp_{(s,a)\sim d_{h}^{\pi^m_k}}\Big[\frac{1}{\lambda_{h}(s,a)}\Big].
\]
Hence the sum over $h$ of the bracket in \eqref{eq:wt_ident} is at most $H^2$ times the braces of \eqref{eq:wt_Lambda} at $\tilde P=\hat P_k$. This is at most $H^2\Lambda^m$, because $\pi^m_k$ is the uniformly tie-broken greedy policy for $\hat P_k\in\mathcal P$. Therefore
\begin{align}
\mexp_{s_1\sim\rho}\big[V^{m,\star}_1(s_1)-V^{m,\pi^m_k}_1(s_1)\big]\le\frac{CH^5\iota\,\Lambda^m}{kM\Delta^m_{\min}}+\frac{CH^4\sqrt S\,\iota}{kM\lambda_{\mathsf{min}}}\qquad(k>K_{th}).\label{eq:wt_reg_k}
\end{align}

\paragraph{Step 5: summation.} We split the episodes as in Step~4. Phase P1 is bounded by \eqref{eq:p1}, hence by $4H\iota/(M\lambda_{\mathsf{min}})+3H$ since $\log(4M^2SAKH/\delta)\le\iota$, and episodes $k>\max\{K_{th},K_0^m\}$ contribute nothing by Lemma~\ref{lemma:tipping_point}, since $K_0^m$ is at least its threshold for $C_0$ large enough. Phase P2 is where the argument departs: we sum \eqref{eq:wt_reg_k} instead of $2H\epsilon_k$, and $\sum_{k=K_{th}+1}^{K_0^m}\frac1k\le\log(1+K_0^m)$ replaces $\sum_k k^{-1/2}\le2\sqrt{K_0^m}$. This gives
\[
\mathsf{Regret}^m_{P2}(K)\le C\log(1+K_0^m)\left[\frac{H^5\iota\,\Lambda^m}{M\Delta^m_{\min}}+\frac{H^4\sqrt S\,\iota}{M\lambda_{\mathsf{min}}}\right].
\]
Summing over $m\in[M]$ and using $K_0^m\le K_0$ gives \eqref{eq:wt_main}.
\end{proof}

Besides the events of Lemmas~\ref{lemma:global_concentration} and~\ref{lemma:shrinking_error}, we use the $\ell_1$ deviation inequality of \citet{weissman2003inequalities}: for $n$ i.i.d.\ samples from a distribution $P$ on $S$ points, $\mprob(\|\hat P_n-P\|_1\ge\varepsilon)\le(2^S-2)e^{-n\varepsilon^2/2}$. We apply it to the first $n$ next states observed at each $(h,s,a)$, as in the proof of Lemma~\ref{lemma:shrinking_error}, with $\varepsilon=\sqrt{2(S\ln2+\iota)/n}$, and take a union bound over $(h,s,a)$ and $n\in[MK]$. With probability at least $1-\delta/4$, for all $k$, $h$ and $(s,a)$ with $N_{k,h}(s,a)\ge1$,
\begin{align}
\big\|\hat P_{k,h}(\cdot|s,a)-P_h(\cdot|s,a)\big\|_1\le\sqrt{\frac{2(S\ln2+\iota)}{N_{k,h}(s,a)}}.\label{eq:wt_l1}
\end{align}
Taking the Bernstein event of Lemma~\ref{lemma:shrinking_error} at level $\delta/4$ instead of $\delta/2$ changes only the constant inside its logarithm. Hence all the events used here hold simultaneously with probability at least $1-\delta$.

\begin{lemma}[One-sided per-pair estimation error]\label{lemma:wt_pair_error}
On the events of Lemmas~\ref{lemma:global_concentration} and~\ref{lemma:shrinking_error} and of \eqref{eq:wt_l1}, for all $k>K_{th}$, $m\in[M]$, $h\in[H]$ and $(s,a)\in\mathcal S_h\times\mathcal A$,
\begin{align}
Q^m_{k,h}(s,a)-Q^{m,\star}_h(s,a)&\le CH\sqrt{\frac{\iota}{kM}}\ \mexp^{\pi^m_k}\Big[\sum_{h'=h}^H\frac{1}{\sqrt{\lambda_{h'}(s_{h'},a_{h'})}}\,\Big|\,s_h=s,a_h=a\Big] +\frac{CH^3\sqrt S\,\iota}{kM\lambda_{\mathsf{min}}},\label{eq:wt_err_up}\\
Q^{m,\star}_h(s,a)-Q^m_{k,h}(s,a)&\le CH\sqrt{\frac{\iota}{kM}}\ \mexp^{\pi^{m,\star}}\Big[\sum_{h'=h}^H\frac{1}{\sqrt{\lambda_{h'}(s_{h'},a_{h'})}}\,\Big|\,s_h=s,a_h=a\Big] +\frac{CH^3\sqrt S\,\iota}{kM\lambda_{\mathsf{min}}}.\label{eq:wt_err_low}
\end{align}
\end{lemma}
}

\section{Proof of Lemmas}

\subsection{Proof of Lemma~\ref{lemma:transition_error_bound}}
By applying the empirical Bernstein inequality \citep{maurer2009empirical} with a union bound over $(h,s,a,m)$ and $N_{k,h}(s,a)\in[MK]$, we can show the upper bound as follows for $N_{k,h}(s,a)\ge2$, using $1/(N_{k,h}(s,a)-1)\le2/N_{k,h}(s,a)$ for its sample-variance normalization and second-order term; for $N_{k,h}(s,a)\le1$, the lemma holds trivially since $B_{k,h}^m(s,a)\ge H$:
\begin{align} \label{eq:transition_error_decomp_1}
\big(\hat{P}_{k,h,s,a} - P_{h,s,a}\big) V_{k,h+1}^m  
&= \big(\hat{P}_{k,h,s,a} - P_{h,s,a}\big) V_{h+1}^{m, \star} + \big(\hat{P}_{k,h,s,a} - P_{h,s,a}\big) (V_{k,h+1}^m - V_{h+1}^{m, \star}) \cr
&\le \sqrt{ \frac{2 \Var_{\hat{P}_{k,h,s,a}}(V_{h+1}^{m, \star}) \iota}{ \max\{N_{k,h}(s,a), 1\}} } + \frac{8 H \iota}{\max\{N_{k,h}(s,a), 1\}} \cr
&\quad + \left| (\hat{P}_{k,h,s,a} - P_{h,s,a})\Delta_{k,h+1}^m \right|  .
\end{align}

\paragraph{Bounding $|(\hat{P}_{k,h,s,a} - P_{h,s,a}) \Delta_{k,h+1}^m|$:} To bound the last term $|(\hat{P}_{k,h,s,a} - P_{h,s,a}) \Delta_{k,h+1}^m|$, we need to consider all possible $\Delta_{k,h+1}^m$ and apply the Bernstein inequality. 
For any fixed $\Delta_{\epsilon}$ that approximates $\Delta_{k,h+1}^m$ by the error of $\epsilon$, by applying the Bernstein inequality and using the fact that $0\le \Delta_{k,h+1}^m \le H$, we can show that
\begin{align}
|(\hat{P}_{k,h,s,a} - P_{h,s,a}) \Delta_{\epsilon}| 
&\le  \min\left \{H, \sqrt{ \frac{2 \Var_{P_{h,s,a}}(\Delta_{\epsilon}) \ln(2/\delta')}{N_{k,h}(s,a)} } + \frac{2 H \log(2/\delta')}{3 N_{k,h}(s,a)} \right\} \cr
&\le \sqrt{ \frac{2 \Var_{P_{h,s,a}}(\Delta_{\epsilon}) \ln(2/\delta')}{\max\{N_{k,h}(s,a),1\}} } + \frac{2 H \log(2/\delta')}{3 \max\{N_{k,h}(s,a),1\}}.
\end{align}
To cover all possible $\Delta_{k,h+1}^m$ with an error of $\epsilon$, we need to consider all possible $\Delta_{\epsilon}$, splitting $[0,H]$ into $\lceil H/\epsilon \rceil$ intervals for all states, incurring covering number $\mathcal{N}(\epsilon) \le \left( 1+ \frac{H}{\epsilon} \right)^S.$ By taking a union bound over $ \mathcal{N}(\epsilon)$, $(h,s,a,m)$, and possible visitation counter $N_{k,h}(s,a) =  N \in [MK]$, we can replace $\delta'$ with $\delta / (7M^2SAKH \mathcal{N}(\epsilon))$ in the above inequality, yielding 
\begin{align}
&|(\hat{P}_{k,h,s,a} - P_{h,s,a}) \Delta_{k,h+1}^m| \cr
&\le |(\hat{P}_{k,h,s,a} - P_{h,s,a}) \Delta_{\epsilon}|  + |(\hat{P}_{k,h,s,a} - P_{h,s,a}) (\Delta_{k,h+1}^m - \Delta_{\epsilon})| \cr
&\le  |(\hat{P}_{k,h,s,a} - P_{h,s,a}) \Delta_{\epsilon}|  + 2 \epsilon \cr
&\le  \sqrt{ \frac{2 \Var_{P_{h,s,a}}(\Delta_{\epsilon}) \log(14 M^2SAKH \mathcal{N}(\epsilon)/\delta)}{ \max\{N_{k,h}(s,a), 1\}} } + \frac{2 H \log(14 M^2SAKH \mathcal{N}(\epsilon)/\delta)}{3 \max\{N_{k,h}(s,a), 1\}} + 2 \epsilon \cr
&\stackrel{(i)}{\le} \sqrt{ \frac{2 H (P \Delta_{k,h+1}^m + \epsilon) \log(14 M^2 SAKH \mathcal{N}(\epsilon)/\delta)}{\max\{N_{k,h}(s,a), 1\}} } + \frac{2 H \log(14 M^2 SAKH \mathcal{N}(\epsilon)/\delta)}{3 \max\{N_{k,h}(s,a), 1\}} +  2 \epsilon \cr
&\stackrel{(ii)}{\le} \sqrt{ \frac{8 S H P \Delta_{k,h+1}^m \iota}{\max\{N_{k,h}(s,a), 1\}} } + \frac{7 H S \iota}{\max\{N_{k,h}(s,a), 1\}} \cr
&\stackrel{(iii)}{\le} \frac{1}{H} P_{h,s,a} \Delta_{k,h+1}^m + \frac{9 S H^2 \iota}{\max\{N_{k,h}(s,a), 1\}} ,
\end{align}
where (i) holds because
\begin{align}
\Var_{P_{h,s,a}}(\Delta_{\epsilon}) 
= P_{h,s,a} \Delta_{\epsilon}^2 - (P_{h,s,a} \Delta_{\epsilon})^2 
\le P_{h,s,a} \Delta_{\epsilon}^2 
\le H P_{h,s,a} \Delta_{\epsilon} 
\le H P_{h,s,a} \Delta_{k,h+1}^m + H \epsilon ,
\end{align}
(ii) holds when choosing very small $\epsilon = 1/(HMK)^2$, since then $\log(14M^2SAKH\mathcal N(\epsilon)/\delta)\le\iota+S\log(1+H^3M^2K^2)\le4S\iota$, and (iii) follows by applying the Young's inequality that $xy \le c x^2 + \frac{y^2}{4c}$ for any $x, y > 0$ and $c > 0$, which yields
\begin{align}
    \sqrt{ \frac{8 S H P_{h,s,a} \Delta_{k,h+1}^m \iota}{\max\{N_{k,h}(s,a), 1\}} }
    \le \frac{1}{H} P_{h,s,a} \Delta_{k,h+1}^m + \frac{2 S H^2 \iota}{\max\{N_{k,h}(s,a), 1\}} 
\end{align}
when  $c = 1/H$, $x=\sqrt{P \Delta_{k,h+1}^m}$, $y=\sqrt{ \frac{8 S H \iota}{\max\{N_{k,h}(s,a), 1\}} } $.

By substituting the above bound into \eqref{eq:transition_error_decomp_1}, we can show that
\begin{align} \label{eq:transition_error_decomp_2}
\big(\hat{P}_{k,h,s,a} - P_{h,s,a}\big) V_{k,h+1}^m  
&\le \sqrt{ \frac{2 \Var_{\hat{P}_{k,h,s,a}}(V_{h+1}^{m, \star}) \iota}{\max\{N_{k,h}(s,a), 1\}} } + \frac{17 S H^2 \iota}{\max\{N_{k,h}(s,a),1\}}   + \frac{1}{H} P_{h,s,a} \Delta_{k,h+1}^m . %
\end{align}

Since $V_{h+1}^{m,\star}=V_{k,h+1}^m-\Delta_{k,h+1}^m$ and $0\le\Delta\le H$,
\[
\Var_{\hat P}(V_{h+1}^{m,\star})\le 2\Var_{\hat P}(V_{k,h+1}^m)+2\Var_{\hat P}(\Delta_{k,h+1}^m)\le 2\Var_{\hat P}(V_{k,h+1}^m)+2H\,\hat P\Delta_{k,h+1}^m ,
\]
hence
\[
\sqrt{\frac{2\Var_{\hat P}(V_{h+1}^{m,\star})\iota}{N}}\le\sqrt{\frac{4\Var_{\hat P}(V_{k,h+1}^m)\iota}{N}}+\sqrt{\frac{4H\iota\,\hat P\Delta_{k,h+1}^m}{N}} .
\]
Using $\hat P\Delta\le 2P\Delta+9SH^2\iota/N$ (from the bound on $|(\hat P-P)\Delta|$ proved above) and $2\sqrt{xy}\le x+y$,
\[
\sqrt{\frac{4H\iota\,\hat P\Delta}{N}}\le\sqrt{\frac{8H\iota\,P\Delta}{N}}+\frac{6\sqrt{S}\,H^{3/2}\iota}{N}\le\frac1H P\Delta+\frac{8SH^2\iota}{N}.
\]

Then, we can rewrite \eqref{eq:transition_error_decomp_2} as
\begin{align}
\big(\hat{P}_{k,h,s,a} - P_{h,s,a}\big) V_{k,h+1}^m  
&\le \underbrace {\sqrt{ \frac{4 \Var_{\hat{P}_{k,h,s,a}}(V_{k,h+1}^{m}) \iota}{\max\{N_{k,h}(s,a), 1\}} } + \frac{25 S H^2 \iota}{\max\{N_{k,h}(s,a),1\}} }_{= B_{k,h}^m(s,a)}   + \frac{2}{H} P_{h,s,a} \Delta_{k,h+1}^m . %
\end{align}

Similarly, we can also show the lower bound as follows:
\begin{align}
\big(\hat{P}_{k,h,s,a} - P_{h,s,a}\big) V_{k,h+1}^m 
&\ge - B_{k,h}^m(s,a) - \frac{2}{H} P_{h,s,a} \Delta_{k,h+1}^m.
\end{align}

\subsection{Proof of Lemma~\ref{lemma:ucbvi_variance_gap}}
For notational simplicity, we omit the subscripts and superscripts in $\hat{P}_{k,h,s,a}$ and $P_{h,s,a}$, and simply denote them as $\hat{P}$ and $P$ respectively. We also denote $V = V_{h+1}^{m, \star}$ for brevity. First, we decompose the variance difference as follows:
\begin{align}
\big| \Var_{\hat{P}}(V) - \Var_{P}(V) \big| &= \big| (\hat{P} V^2 - (\hat{P} V)^2) - (P V^2 - (P V)^2) \big| \cr
&= \big| (\hat{P} V^2 - P V^2) + ((P V)^2 - (\hat{P} V)^2) \big| \cr
&\le \big| (\hat{P} - P) V^2 \big| +  \big| (P  + \hat{P} )V \cdot (\hat{P} - P) V \big|  \cr
&\le \big| (\hat{P} - P) V^2 \big|  + 2H \big| (\hat{P} - P) V \big| 
\end{align}
Then, applying the Bernstein inequality to both terms, we can show that
\begin{align}
\big| (\hat{P} - P) V^2 \big| &\le \sqrt{\frac{2 \Var_P(V^2) \iota}{N}} + \frac{2 H^2 \iota}{3 N} \quad (V^2 \in [0, H^2]) \cr
&\le \sqrt{\frac{8 H^2 \Var_P(V) \iota}{N}} + \frac{2 H^2 \iota}{3 N} , \\
\big| (\hat{P} - P) V \big| &\le \sqrt{\frac{2 \Var_P(V) \iota}{N}} + \frac{2H \iota}{3N} ,
\end{align}
using the fact that
\begin{align}
\Var_P(V^2) &= \frac{1}{2} \mexp_{x, y} \left[ (V(x)^2 - V(y)^2)^2 \right] \cr
&= \frac{1}{2} \mexp_{x, y} \left[ (V(x) - V(y))^2 (V(x) + V(y))^2 \right] \le  4H^2 \Var_P(V).
\end{align}
Combining the above bounds, we have:
\begin{align}
\big| \Var_{\hat{P}}(V) - \Var_{P}(V) \big| 
&\le \min \Big \{H^2, \sqrt{\frac{32 H^2 \Var_P(V) \iota}{N}} + \frac{2 H^2 \iota}{ N} \Big \}.
\end{align}
Applying the Young's inequality,
\begin{align}
\sqrt{\frac{32 H^2 \text{Var}_P(V) \iota}{N}} 
= 2 \cdot \sqrt{\text{Var}_P(V)} \cdot \sqrt{\frac{8 H^2 \iota}{N}} 
\le \text{Var}_P(V) + \frac{8 H^2 \iota}{N}
\end{align}

Thus, we have
\begin{align}
\big| \Var_{\hat{P}}(V) - \Var_{P}(V) \big| \le \Var_P(V) + \frac{10 H^2 \iota}{\max\{N,1\}}.
\end{align}

\subsection{Proof of Lemma~\ref{lemma:global_concentration}}
We clarify the relationship between the estimated dynamics 
$\hat{P}_{\tau,h}$ and the coverage guarantee. While each agent $m$ 
computes its greedy policy $\pi_{\tau}^m$ using the estimated transition 
kernel $\hat{P}_{\tau,h} \in \mathcal{F}_{\tau-1}$, the marginal visitation 
probability $d_h^{\pi_{\tau}^m}(s, a)$ is always evaluated under the 
\textit{true} transition kernel $P_h$, not under $\hat{P}_{\tau,h}$. 
Since $\hat P_{\tau,h}(\cdot|s,a)$ belongs to the set of all valid transition kernels $\mathcal{P}$,
Assumption~\ref{assumption:expfree_global_coverage} applies directly
with $\tilde{P} = \hat{P}_{\tau,h}$: regardless of how inaccurate 
$\hat{P}_{\tau,h}$ may be relative to the true dynamics $P_h$, the 
diversity of the agents' reward functions $\{r^m\}_{m=1}^M$ structurally 
ensures that the resulting greedy policies $\{\pi_{\tau}^m\}_{m=1}^M$ 
collectively cover the state-action space under the true environment, 
yielding $\sum_{m=1}^M d_h^{\pi_{\tau}^m}(s, a) \ge M\lambda_{\mathsf{min}}$ for
every episode $\tau$.
Here $\pi^m_\tau$ is the greedy policy with uniform tie-breaking, which is exactly the policy $\tilde\pi^m$ of Assumption~\ref{assumption:expfree_global_coverage} for $\tilde P=\hat P_{\tau,h}$.

Let $\mathcal F_0$ be the trivial $\sigma$-field and, for $\tau\ge1$, let
\[
\mathcal F_\tau=\sigma\Big(\big\{s^m_{i,h}:m\in[M],\,i\le\tau,\,h\in[H+1]\big\}\cup\big\{a^m_{i,h}:m\in[M],\,i\le\tau,\,h\in[H]\big\}\Big)
\]
be the $\sigma$-field generated by the states and actions of all agents in episodes $1,\dots,\tau$.
At the beginning of episode $\tau$, the central server broadcasts the empirical transition kernel $\hat{P}_{\tau,h}$, which is fully determined by $\mathcal{F}_{\tau-1}$.
Conditioned on the filtration $\mathcal{F}_{\tau-1}$, each agent $m \in [M]$ computes its purely greedy policy $\pi_{\tau,h}^m$ based on its local reward $r_h^m$ and the shared estimated dynamics $\hat{P}_{\tau,h}$. Let $d_h^{\pi_{\tau}^m}(s, a)$ be the exact marginal probability of visiting $(s,a)$ at step $h$ under this executed policy $\pi_\tau^m$.

{Fix $(h,s,a)$ and order the indicators $\mathbb{I}(s^m_{\tau,h}=s,a^m_{\tau,h}=a)$ lexicographically in $(\tau,m)$, refining $(\mathcal F_\tau)$ by revealing the agents' episode-$\tau$ trajectories one at a time. Given $\mathcal F_{\tau-1}$, the episode-$\tau$ trajectories of the agents are independent, since the agents interact with independent copies of the environment, and agent $m$'s trajectory follows $\pi^m_\tau$ in $P$ from $s_1\sim\rho$. Hence, given everything preceding position $(\tau,m)$ in this order, the indicator at $(\tau,m)$ equals one with probability $d_h^{\pi^m_\tau}(s,a)$, which is $\mathcal F_{\tau-1}$-measurable since $\pi^m_\tau$ is determined by $r^m$ and $\hat P_\tau$. The lower-tail bound for adapted Bernoulli sequences of \citet[Lemma~F.4]{dann2017unifying} then gives, with probability at least $1-\delta'$, simultaneously for all $k\ge1$,
\begin{align}
N_{k+1,h}(s,a)\ge\frac12\sum_{\tau=1}^k\sum_{m=1}^Md_h^{\pi^m_\tau}(s,a)-\ln\frac{1}{\delta'}.\label{eq:count_ville}
\end{align}
We take a union bound over $(h,s,a)$ with $\delta'=\delta/(2SAH)$.}

As shown above, Assumption~\ref{assumption:expfree_global_coverage} applies with $\tilde P=\hat P_{\tau,h}\in\mathcal P$. Thus, for every episode $\tau\ge2$ (in episode~1 all actions tie under the zero initialization, so $\pi^m_1$ is uniform and we only use $d_h^{\pi^m_1}\ge0$):
\begin{align}
    \sum_{m=1}^M d_h^{\pi_{\tau}^m}(s, a) \ge M \lambda_{\mathsf{min}}.
\end{align}

{Substituting this lower bound into the concentration inequality above, we obtain, for all $k\ge1$, $h\in[H]$ and $(s,a)$,
\begin{align}
N_{k+1,h}(s,a)\ge\frac12(k-1)M\lambda_{\mathsf{min}}-\ln\frac{2SAH}{\delta}.
\end{align}
For $k\ge K_{th}$, we have $\ln(2SAH/\delta)\le\log(4M^2SAKH/\delta)\le\frac14(k-2)M\lambda_{\mathsf{min}}$, and hence $N_{k+1,h}(s,a)\ge\frac12(k-1)M\lambda_{\mathsf{min}}-\frac14(k-2)M\lambda_{\mathsf{min}}\ge\frac14kM\lambda_{\mathsf{min}}$. This completes the proof.}

\subsection{Proof of Lemma~\ref{lemma:shrinking_error}}
We bound the absolute estimation error of the Q-value by backward induction on $h$. For the base case $h = H+1$, the error is $0$.

For any step $h \in [H]$, taking the absolute difference between the empirical Q-value update and the true optimal Bellman equation, we apply H\"older's inequality:
\begin{align} \label{eq:bellman_decomp}
    \left| Q_{k,h}^m(s, a) - Q_h^{m, \star}(s, a) \right| 
    &\le \left| \langle \hat{P}_{k,h}(s,a) - P_h(s,a), V_{h+1}^{m, \star} \rangle \right| + \left| \langle \hat{P}_{k,h}(s,a), V_{k,h+1}^m - V_{h+1}^{m, \star} \rangle \right|
\end{align}

Let $s_{1,h+1},s_{2,h+1},\dots\overset{\text{i.i.d.}}{\sim}P_h(\cdot|s,a)$ be the successive next states observed at $(h,s,a)$, so that $\langle\hat P_{k,h}(s,a),V\rangle$ equals $\frac1n\sum_{i\le n}V(s_{i,h+1})$ on $\{N_{k,h}(s,a)=n\}$. For each fixed $n\in[MK]$, Bernstein's inequality gives, with probability at least $1-\delta'$,
\begin{align}
    \big|\langle \hat{P}_{k,h} - P_h, V_{h+1}^{m, \star} \rangle\big| 
    &\le \sqrt{\frac{2 \Var_{P_h}(V_{h+1}^{m, \star}) \log(\frac{2}{\delta'}) }{ N_{k,h}(s, a)}} + \frac{2 \|V_{h+1}^{m, \star}\|_{\infty} \log(\frac{2}{\delta'})}{3 N_{k,h}(s, a)} \label{eq:bern_pair}
\end{align}

For $k>K_{th}$, Lemma~\ref{lemma:global_concentration} at $k-1\ge K_{th}$ gives $N_{k,h}(s,a)\ge\frac14(k-1)M\lambda_{\mathsf{min}}\ge\frac18kM\lambda_{\mathsf{min}}$, and $N_{k,h}(s,a)\ge1$ since $\frac14K_{th}M\lambda_{\mathsf{min}}\ge\log(4M^2SAKH/\delta)>1$. Substituting this lower bound and $\|V_{h+1}^{m, \star}\|_{\infty} \le H$, we have:
\begin{align}
    \langle \hat{P}_{k,h} - P_h, V_{h+1}^{m, \star} \rangle 
    \le \mathcal{O} \left( \sqrt{\frac{ \Var_{P_h}(V_{h+1}^{m, \star}) \log(\frac{2}{\delta'}) }{ M \lambda_{\mathsf{min}} k}} + \frac{H \log(\frac{2}{\delta'})}{M \lambda_{\mathsf{min}} k} \right )
\end{align}

The second term can be recursively unpacked as the maximum estimation error of the value function at step $h+1$. As the errors propagate backward until reaching the last step $H$, the maximum error accumulated over $H$ steps is bounded by: %
\begin{align}
    \max_{s,a} \left| Q_{k,h}^m(s, a) - Q_h^{m, \star}(s, a) \right| 
    &\le  \mathcal{O}  \left( \sqrt{\frac{ \log(\frac{2}{\delta'}) }{ M \lambda_{\mathsf{min}} k}} \sum_{h'=h}^H  \sqrt{ \Var_{P_{h'}}(V_{h'+1}^{m, \star}) } +  \frac{H^2 \log(\frac{2}{\delta'})}{M \lambda_{\mathsf{min}} k}  \right ) \cr
    &\le  \mathcal{O}  \left( \sqrt{\frac{ \log(\frac{2}{\delta'}) }{ M \lambda_{\mathsf{min}} k}} H^2 +  \frac{H^2 \log(\frac{2}{\delta'})}{M \lambda_{\mathsf{min}} k}  \right ) 
\end{align}
Here, we use the fact that the variance of the optimal value function is bounded by $H^2$, implying $\sqrt{\Var_{P_{h'}}(V_{h'+1}^{m, \star})} \le H$. Summing this worst-case bound over $H$ steps yields $H^2$, which is then absorbed into the square root as $H^4$.

We require this bound to hold uniformly for all $k \in [K]$, $m \in [M]$, $h \in [H]$, $s \in \mathcal{S}_h$, and $a \in \mathcal{A}$. Taking a union bound over $(m,h,s,a)$ and $n\in[MK]$, which covers every $k\in[K]$, and allocating the remaining failure probability $\delta/2$, we set $\delta'=\delta/(2M^2KHSA)$, so that $\log(2/\delta')=\log(4M^2KHSA/\delta)\le2\log(4KMHSA/\delta)$. Then,
\begin{align}
    \max_{s,a} \left| Q_{k,h}^m(s, a) - Q_h^{m, \star}(s, a) \right| 
    &\le \mathcal{O}  \left( \sqrt{\frac{ H^4 \log(4 K M H S A / \delta) }{ M \lambda_{\mathsf{min}} k}} + \frac{H^2 \log(4 K M H S A / \delta)}{M \lambda_{\mathsf{min}} k} \right )
\end{align}

This completes the proof.

\subsection{Proof of Lemma~\ref{lemma:tipping_point}}
Let $\epsilon_k = \max_{s,a,h} | Q_{k,h}^m(s, a) - Q_h^{m, \star}(s, a) |$ be the maximum absolute estimation error at episode $k$. By Lemma \ref{lemma:shrinking_error}, we have $\epsilon_k \le \mathcal{O}\left(\sqrt{\frac{H^4 \log(K M H S A / \delta)}{k M \lambda_{\mathsf{min}}}}\right)$. Here the second-order term of Lemma~\ref{lemma:shrinking_error} is dropped because that term is at most the first-order term if $kM\lambda_{\mathsf{min}}\ge\log(4M^2KHSA/\delta)$, which holds for $k>K_{th}$ since $kM\lambda_{\mathsf{min}}>K_{th}M\lambda_{\mathsf{min}}\ge4\log(4M^2SAKH/\delta)$.

Because the agent acts greedily without any UCB bonuses, it will strictly prefer the optimal action $a^\star$ over any suboptimal action $a'$ if:
\begin{align}
    Q_{k,h}^m(s, a^\star) > Q_{k,h}^m(s, a')
\end{align}
Using the absolute error bound, the estimated values satisfy $Q_{k,h}^m(s, a^\star) \ge Q_h^{m, \star}(s, a^\star) - \epsilon_k$ and $Q_{k,h}^m(s, a') \le Q_h^{m, \star}(s, a') + \epsilon_k$. Therefore, the optimal action is guaranteed to be chosen if:
\begin{align}
    Q_h^{m, \star}(s, a^\star) - \epsilon_k > Q_h^{m, \star}(s, a') + \epsilon_k \implies 2 \epsilon_k < Q_h^{m, \star}(s, a^\star) - Q_h^{m, \star}(s, a') = \Delta_h^m(s, a')
\end{align}
To ensure this holds for all suboptimal actions, we require $2 \epsilon_k < \Delta_{\min}^m$. Substituting the upper bound of $\epsilon_k$ and solving for $k$, we find that the condition holds for all $k > K_0^m$. For $k > \max(K_{th}, K_0^m)$, no suboptimal actions are ever chosen.

\subsection{Proof of Lemma~\ref{lemma:wt_pair_error}}
{Fix $k>K_{th}$ and $m$. The proof of Lemma~\ref{lemma:global_concentration} applies to each fixed $(h,s,a)$. Its coverage step $\frac1M\sum_md_h^{\pi^m_\tau}(s,a)\ge\lambda_{\mathsf{min}}$ can be replaced by $\frac1M\sum_md_h^{\pi^m_\tau}(s,a)\ge\lambda_h(s,a)$, which holds by the definition of $\lambda_h$ because $\pi^m_\tau$ is the uniformly tie-broken greedy policy for $\hat P_\tau\in\mathcal P$ for $\tau\ge2$. Combining this with the count argument in the proof of Lemma~\ref{lemma:shrinking_error}, where $K_{th}$ is unchanged because $\lambda_h(s,a)\ge\lambda_{\mathsf{min}}$, gives, for all $(h,s,a)$,
\begin{align}
N_{k,h}(s,a)\ge1,\qquad N_{k,h}(s,a)\ge\frac18kM\lambda_h(s,a).\label{eq:wt_count_lin}
\end{align}

Since $\hat P_{k,h}(s,a)$ is a probability vector, the identity behind \eqref{eq:bellman_decomp} reads $Q^m_{k,h}(s,a)-Q^{m,\star}_h(s,a)=\langle\hat P_{k,h}(s,a)-P_h(s,a),V^{m,\star}_{h+1}\rangle+\langle\hat P_{k,h}(s,a),V^m_{k,h+1}-V^{m,\star}_{h+1}\rangle$.
For every $s'$,
\begin{align}
Q^m_{k,h+1}(s',\pi^{m,\star}_{h+1}(s'))-Q^{m,\star}_{h+1}(s',\pi^{m,\star}_{h+1}(s'))&\le V^m_{k,h+1}(s')-V^{m,\star}_{h+1}(s')\nonumber\\
&\le\sum_{a'}\pi^m_{k,h+1}(a'|s')\big[Q^m_{k,h+1}(s',a')-Q^{m,\star}_{h+1}(s',a')\big].\label{eq:wt_sandwich}
\end{align}
The lower bound holds because $V^m_{k,h+1}(s')=\max_{a'}Q^m_{k,h+1}(s',a')$, which is at least $Q^m_{k,h+1}(s',\pi^{m,\star}_{h+1}(s'))$, and because $V^{m,\star}_{h+1}(s')=Q^{m,\star}_{h+1}(s',\pi^{m,\star}_{h+1}(s'))$. For the upper bound, $\pi^m_{k,h+1}(\cdot|s')$ is supported on $\argmax_{a'}Q^m_{k,h+1}(s',a')$, so $V^m_{k,h+1}(s')=\sum_{a'}\pi^m_{k,h+1}(a'|s')Q^m_{k,h+1}(s',a')$, while $V^{m,\star}_{h+1}(s')\ge\sum_{a'}\pi^m_{k,h+1}(a'|s')Q^{m,\star}_{h+1}(s',a')$.

The substitution in the proof of Lemma~\ref{lemma:shrinking_error}, with the per-pair count \eqref{eq:wt_count_lin} in place of the count of Lemma~\ref{lemma:global_concentration}, gives $|\langle\hat P_{k,h}(s,a)-P_h(s,a),V^{m,\star}_{h+1}\rangle|\le CH\sqrt{\iota/(kM\lambda_h(s,a))}+CH\iota/(kM\lambda_{\mathsf{min}})$. In the same way, \eqref{eq:wt_l1} gives $\|\hat P_{k,h}(s,a)-P_h(s,a)\|_1\le C\sqrt{(S+\iota)/(kM\lambda_{\mathsf{min}})}$. For the maximal error $\epsilon_k$ of Lemma~\ref{lemma:tipping_point}, Lemma~\ref{lemma:shrinking_error} gives $\epsilon_k\le C\sqrt{H^4\iota/(kM\lambda_{\mathsf{min}})}$ for $k>K_{th}$, since its second-order term is dominated there, because $kM\lambda_{\mathsf{min}}>K_{th}M\lambda_{\mathsf{min}}\ge4\log(4M^2SAKH/\delta)\ge\iota$. Combining these bounds and using $\sqrt{(S+\iota)\iota}\le2\sqrt S\,\iota$ and $H\iota\le H^2\sqrt S\,\iota$,
\begin{align}
\big|\big\langle\hat P_{k,h}(s,a)-P_h(s,a),V^{m,\star}_{h+1}\big\rangle\big|+\big\|\hat P_{k,h}(s,a)-P_h(s,a)\big\|_1\epsilon_k&\le CH\sqrt{\frac{\iota}{kM\lambda_h(s,a)}} +\frac{CH^2\sqrt S\,\iota}{kM\lambda_{\mathsf{min}}}.\label{eq:wt_step}
\end{align}

Start from the identity above and apply the upper half of \eqref{eq:wt_sandwich}, which is allowed because $\hat P_{k,h}(s,a)\ge0$. Then write $\hat P_{k,h}(s,a)=P_h(s,a)+(\hat P_{k,h}(s,a)-P_h(s,a))$, and bound the second part by $\|\hat P_{k,h}(s,a)-P_h(s,a)\|_1\epsilon_k$, since $|Q^m_{k,h+1}-Q^{m,\star}_{h+1}|\le\epsilon_k$. With \eqref{eq:wt_step}, this gives
\begin{align}
Q^m_{k,h}(s,a)-Q^{m,\star}_h(s,a)&\le CH\sqrt{\frac{\iota}{kM\lambda_h(s,a)}}+\frac{CH^2\sqrt S\,\iota}{kM\lambda_{\mathsf{min}}}\nonumber\\
&\quad+\sum_{s'}P_h(s'|s,a)\sum_{a'}\pi^m_{k,h+1}(a'|s')\big[Q^m_{k,h+1}(s',a')-Q^{m,\star}_{h+1}(s',a')\big].\label{eq:wt_rec_up}
\end{align}
In the same way, the lower half of \eqref{eq:wt_sandwich} gives
\begin{align}
Q^{m,\star}_h(s,a)-Q^m_{k,h}(s,a)&\le CH\sqrt{\frac{\iota}{kM\lambda_h(s,a)}}+\frac{CH^2\sqrt S\,\iota}{kM\lambda_{\mathsf{min}}}\nonumber\\
&\quad+\sum_{s'}P_h(s'|s,a)\big[Q^{m,\star}_{h+1}(s',\pi^{m,\star}_{h+1}(s'))-Q^m_{k,h+1}(s',\pi^{m,\star}_{h+1}(s'))\big].\label{eq:wt_rec_low}
\end{align}
For $h=H$, the last sums are $0$, because $V^m_{k,H+1}=V^{m,\star}_{H+1}=0$.

We prove \eqref{eq:wt_err_up} with $CH^3$ in the second-order term replaced by $(H-h+1)CH^2$, by backward induction on $h$. For $h=H$, the right-hand side of \eqref{eq:wt_rec_up} is exactly the claim, because $\mexp^{\pi^m_k}[\lambda_H(s_H,a_H)^{-1/2}\mid s_H=s,a_H=a]=\lambda_H(s,a)^{-1/2}$. Suppose the claim holds at $h+1$ for all $(s',a')$, and substitute it into the last sum of \eqref{eq:wt_rec_up}. The weights $P_h(s'|s,a)\pi^m_{k,h+1}(a'|s')$ are nonnegative and sum to one, and, by the tower property,
\begin{align*}
&\sum_{s'}P_h(s'|s,a)\sum_{a'}\pi^m_{k,h+1}(a'|s')\,\mexp^{\pi^m_k}\Big[\sum_{h'=h+1}^H\frac{1}{\sqrt{\lambda_{h'}(s_{h'},a_{h'})}}\,\Big|\,s_{h+1}=s',a_{h+1}=a'\Big]\\
&\qquad=\mexp^{\pi^m_k}\Big[\sum_{h'=h+1}^H\frac{1}{\sqrt{\lambda_{h'}(s_{h'},a_{h'})}}\,\Big|\,s_h=s,a_h=a\Big].
\end{align*}
Adding the step-$h$ term $CH\sqrt{\iota/(kM)}\,\lambda_h(s,a)^{-1/2}$ completes the expectation from $h'=h$, and the second-order terms add up to $(H-h)CH^2+CH^2=(H-h+1)CH^2\le CH^3$. This proves \eqref{eq:wt_err_up}. The proof of \eqref{eq:wt_err_low} is identical, using \eqref{eq:wt_rec_low} with $\pi^{m,\star}$ in place of $\pi^m_k$.
}

\section{Experimental details}
\label{sec:exp-details}
We use synthetic finite-horizon tabular MDPs with heterogeneous rewards. We draw $P_h(\cdot \mid s, a) \sim \text{Dirichlet}(\alpha \cdot \mathbf{1})$, $\rho$ is uniform, and small $\alpha$ yields concentrated dynamics that trap greedy policies.
Each agent is assigned an independently drawn random subset of $\lceil \kappa \cdot SAH \rceil$ preferred $(h, s, a)$ cells with reward $1$ (and $0$ elsewhere); small $\kappa$ makes the rewards sparse, so myopic behaviour rarely reaches them. Unless stated otherwise we use $S = 10$, $A= 4$, $H = 5$, $\alpha = 0.05$, and $\kappa = 0.015$ (three reward cells out of $200$); when sweeping $M$ we share a base draw of $M_{\max}$ rewards and take the first $M$ so only the federation size changes. Results average over multiple MDP draws; shaded bands show one standard deviation. We set $\widehat{P}_h(\cdot \mid s, a)$ to uniform at unvisited $(h, s, a)$ to keep it a valid kernel, and scale the theoretical UCB bonus by a constant $c_{\text{bonus}}$ tuned separately for each UCB algorithm (\pfucbvi{} and Fed-UCBVI) on the same held-out MDP over $\{10^{-5}, \dots, 10^{-1}\}$; the selected value is $10^{-5}$ for both.

{\paragraph{Comparison with a consensus baseline.}
We also compare with Fed-UCBVI~\citep{labbi2024federated}, which learns a single consensus policy maximizing the average value across agents, in the setup of Section~\ref{sec:setup}. Table~\ref{tab:feducbvi} reports the per-agent regret at $K=1500$, obtained from a separate set of runs with different random seeds, so the values differ from those in Figure~\ref{fig:regret}. With $M=1$, where there is no reward heterogeneity, \pfucbvi{} and Fed-UCBVI perform comparably. With $M=4,16$ agents of distinct rewards, the per-agent regret of Fed-UCBVI grows with $M$, since the consensus policy can be suboptimal for individual agents, whereas that of \pfucbvi{} and \pfefvi{} decreases with $M$.

\begin{table}[h]
\centering
\caption{Per-agent regret at $K=1500$ (mean $\pm$ std over $8$ MDP draws).}
\label{tab:feducbvi}
\begin{tabular}{cl cc}
\toprule
$M$ & Algorithm & Cumulative regret / agent & Instantaneous regret / agent \\
\midrule
\multirow{3}{*}{1}  & \pfucbvi  & $87.8\pm8.8$     & $0.011\pm0.007$ \\
                    & \pfefvi   & $205.4\pm130.2$  & $0.113\pm0.083$ \\
                    & Fed-UCBVI & $107.0\pm28.9$   & $0.016\pm0.011$ \\
\midrule
\multirow{3}{*}{4}  & \pfucbvi  & $25.0\pm5.3$     & $0.002\pm0.001$ \\
                    & \pfefvi   & $33.1\pm16.5$    & $0.010\pm0.009$ \\
                    & Fed-UCBVI & $744.9\pm132.1$  & $0.481\pm0.089$ \\
\midrule
\multirow{3}{*}{16} & \pfucbvi  & $7.3\pm1.1$      & $0.000\pm0.000$ \\
                    & \pfefvi   & $7.7\pm1.3$      & $0.001\pm0.001$ \\
                    & Fed-UCBVI & $1162.7\pm130.9$ & $0.771\pm0.087$ \\
\bottomrule
\end{tabular}
\end{table}
}

\end{document}